\documentclass[conference]{IEEEtran}% \documentclass[letterpaper, 10pt, conference]{ieeeconf}  % Comment this line out if you need a4paper

\IEEEoverridecommandlockouts                              % This command is only needed if 
\usepackage{color}
\usepackage{colortbl}
\usepackage{tabularx}
\usepackage{tcolorbox}
\usepackage{courier}
\usepackage{graphbox}
\usepackage{booktabs}
\usepackage{amsmath,amsfonts}
\usepackage{graphicx} 
\usepackage{subcaption}
\usepackage{tabularray}
\usepackage{url}
\usepackage{etoolbox}
\usepackage{wasysym}
\usepackage[export]{adjustbox}
\usepackage{tikz}
\usepackage{bm}
\usepackage{cite}
\usepackage{academicons}
\usepackage{multirow}
\usepackage{multicol}
\usepackage{algorithm}
\usepackage{algorithmic}
\usepackage{pifont}
\usepackage[font=footnotesize]{caption}
\usepackage{tocloft} % 导言区加载一次即可
\usepackage{geometry}
\usepackage{listings}

\definecolor{orcidlogocol}{HTML}{A6CE39}
\definecolor{lime}{HTML}{A6CE39}

\makeatletter
\let\NAT@parse\undefined
\makeatother

\usepackage{hyperref}
\hypersetup{
    linkcolor=black,
    citecolor=black,
    urlcolor=blue
}
\usepackage[capitalize]{cleveref}
\crefname{figure}{Fig.}{Figs.}
\Crefname{figure}{Figure}{Figures}
\crefname{section}{Sec.}{Secs.}
\Crefname{section}{Section}{Sections}
\crefname{table}{Tab.}{Tabs.}
\Crefname{table}{Table}{Tables}
\crefname{equation}{Eq.}{Eqs.}
\Crefname{equation}{Equation}{Equations}

\makeatletter
\def\@maketitle{%
  \newpage
  \null
\vspace*{2.5em} % <<< 加这一行控制整体下移，推荐 2~3em 可自行调

  \begin{center}%
    {\large \bf \@title \par}%
    \vskip 0.8em%
    {\large
      \lineskip .5em%
      \begin{tabular}[t]{c}%
        \@author
      \end{tabular}\par
    }%
    \vspace{25pt}%
    {\fontsize{10pt}{20pt}\selectfont\ttfamily
      \href{https://drivevla.github.io}{\textcolor[HTML]{6A5ACD}{https://drivevla.github.io}}%
    }%

        \vspace{10pt}%

    \setcounter{figure}{0}% 确保从 Fig.1 开始
    \includegraphics[width=1.0\linewidth]{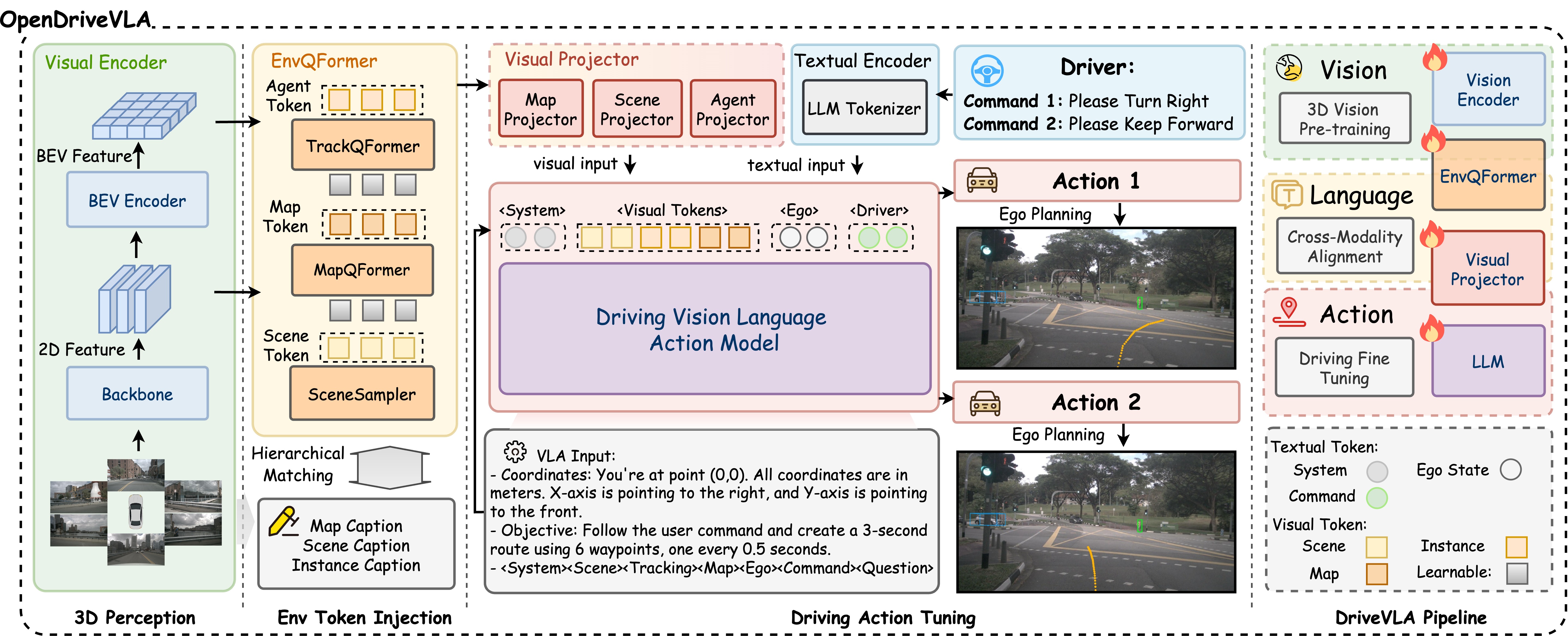}%

    \captionsetup{type=figure}
    \caption{\small
      OpenDriveVLA leverages open-source pre-trained large vision-language-action models to generate driving actions conditioned on 3D environmental perception, ego-vehicle states, and driver commands. It achieves strong performance in both open-loop planning and driving-related question answering, demonstrating its proficiency in scene understanding and driving action tuning.
    }%
    \label{fig:title_figure}
  \end{center}%
  \par
}
\makeatother

\title{\LARGE \bf OpenDriveVLA: Towards End-to-end Autonomous Driving with Large Vision Language Action Model}

\author{
\IEEEauthorblockN{
Xingcheng Zhou\textsuperscript{1†},\thanks{\textsuperscript{†}Corresponding author: \texttt{xingcheng.zhou@tum.de}}
Xuyuan Han\textsuperscript{1},
Feng     Yang\textsuperscript{1},
Yunpu Ma\textsuperscript{2},
Volker Tresp\textsuperscript{2},
Alois Knoll\textsuperscript{1}
}
\vspace{-1em} % 缩小作者和单位之间的间距

\IEEEauthorblockA{
\textsuperscript{1}Technical University of Munich \hspace{0.2em}
\textsuperscript{2}Ludwig Maximilian University of Munich\\
}
}
\begin{document}

%\begin{sloppypar}
\maketitle
% \thispagestyle{empty}
% \pagestyle{empty}
%%%%%%%%%%%%%%%%%%%%%%%%%%%%%%%%%%%%%%%%%%%%%%%%%%%%%%%%%%%%%%%%%%%%%%%%%%%%%%%%
%  figure plan
%  [x] Figure 1 Student Teacher pipeline
%  [x] Figure 2 TUM-Traffic synthetic Dataset statistics show
%  [x] Figure 3 Performance Comparision 

% table plan
%  [x] ablation: yolo compare
%  [x] ablation: with/without DA framework
%  [x] compare with existing: infradet3d
%  [x] with TUM synthetic 3k samples vs 20 samples 

%  [] cite thesis of map.. 

%%%%%%%%%%%%%%%%%%%%%%%%%%%%%%%%%%%%%%%%%%%%%%%%%%%%%%%%%%%%%%%%%%%%%%%%%%%%%%%%

% \begin{figure*}[H]
%     \centering
%    {\includegraphics[width=\linewidth, height=100pt]{figure/Frame_work_v7.jpg}}% define the image
%     \caption{Workflow of DA-RM3D Framework.}
%     \label{fig:odvsor}
% \end{figure*}

% https://app.diagrams.net/#G1KDS1JGpg2x1D42rrEl5xI3TE2LVLJbKb#%7B%22pageId%22%3A%226AU5PCprTE4PCitmrw3E%22%7D

\begin{abstract}
We present OpenDriveVLA, a Vision-Language Action (VLA) model designed for end-to-end autonomous driving, built upon open-source large language models. OpenDriveVLA generates spatially-grounded driving actions by leveraging multimodal inputs, including both 2D and 3D instance-aware visual representations, ego vehicle states, and language commands. To bridge the modality gap between driving visual representations and language embeddings, we introduce a hierarchical vision-language alignment process, projecting both 2D and 3D structured visual tokens into a unified semantic space. Furthermore, we incorporate structured agent–environment–ego interaction modeling into the autoregressive decoding process, enabling the model to capture fine-grained spatial dependencies and behavior-aware dynamics critical for reliable trajectory planning. Extensive experiments on the nuScenes dataset demonstrate that OpenDriveVLA achieves state-of-the-art results across open-loop trajectory planning and driving-related question-answering tasks. Qualitative analyses further illustrate its superior capability to follow high-level driving commands and generate trajectories under challenging scenarios, highlighting its potential for next-generation end-to-end autonomous driving.
\end{abstract}
\section{Introduction}
% [x] figure right append [Action] 

End-to-end learning frameworks have emerged as a promising paradigm in autonomous driving, enabling perception, prediction, and planning to be jointly optimized within a unified neural network \cite{vlmadsurvey}. They learn policies directly from sensor inputs and generalize well across varied scenarios. Despite notable progress, existing approaches still face critical challenges, including limited long-tail generalization, poor complex semantics understanding, and rigid task reasoning \cite{chen2023e2esurvey}. Meanwhile, large language models (LLMs) and vision-language models (VLMs) exhibit strong in-context reasoning, commonsense understanding, and zero-shot generalization abilities. These capabilities are promising for driving, where robust scene understanding is crucial \cite{datasetsurvey,zhou2024gpt4vtrafficassistantindepth}. However, directly leveraging existing VLMs for autonomous driving poses fundamental challenges. Firstly, current VLMs are predominantly optimized for static, 2D image-language tasks, leading to poor spatial reasoning performance in dynamic 3D driving environments \cite{siglip}. Besides, instance-agnostic VLMs \cite{liu2024surveyhallucinationlargevisionlanguage1} are prone to hallucinations, often yielding incorrect yet overconfident outputs, posing safety risks in autonomous driving. Motivated by these limitations, our work answers a central question: \textbf{How can we harness the emergent capabilities of large VLMs to produce safe spatially-grounded driving actions in dynamic 3D environments, while balancing inference speed and planning effectiveness?} 
% Specifically, we aim to repurpose vision-language models, originally built for static 2D image-language understanding, toward real-time 3D-aware end-to-end driving.

% Unlike prior works that apply VLMs only to 2D static perception or offline scene understanding, OpenDriveVLA generates reliable driving trajectories conditioned on multimodal inputs, including instance-aware environmental perception, ego state, and driver commands. 

To enhance spatial-awareness and safety in LLM-based vision-language action model, we introduce two key designs. First, we structure the driving environment using instance-aware, hierarchical 2D and 3D visual representations to reduce the risk of instance hallucinations. Second, we incorporate agent–environment–ego interaction modeling, which is originally explicitly modeled in traditional end-to-end driving systems, as an auxiliary objective into the autoregressive LLM training pipeline. It enables the model to internalize physical feasibility and dynamic multi-agent interactions, improving robustness in safety-critical scenarios.

% We present OpenDriveVLA, a novel vision-language action model designed for end-to-end autonomous driving. 
Built upon open-source large language models, OpenDriveVLA tightly integrates spatially-grounded multimodal reasoning and driving trajectory generation within a unified autoregressive framework. Unlike prior VLM-based methods, OpenDriveVLA leverages structured 2D and 3D instance-aware representations, ego vehicle states, and high-level commands to directly produce reliable driving actions. Extensive experiments on nuScenes benchmark demonstrate that OpenDriveVLA achieves state-of-the-art performance in both open-loop planning and vision-language reasoning tasks. Our key contributions are:

% Extensive experiments on nuScenes dataset show that OpenDriveVLA establishes new state-of-the-art results in both open-loop planning and driving-related question answering, consistently outperforming prior autonomous driving approaches. 

% \begin{itemize}
%     \item We present OpenDriveVLA, the first vision-language action model for 3D end-to-end autonomous driving, capable of generating reliable driving trajectories from rich multimodal inputs, including 3D perception, ego state, and language commands.

%     \item We introduce a hierarchical vision-language feature alignment module, projecting structured 2D and 3D visual tokens into a unified semantic embedding space to facilitate language-guided trajectory generation.
    
%     \item We design an agent-env-ego interaction process to capture interactions among the ego vehicle, dynamic agents, and static map elements, significantly enhancing motion forecasting accuracy and trajectory reliability in complex traffic scenarios.

% \end{itemize}

 \begin{itemize}
    \item We present OpenDriveVLA, a 3D vision-language action model for end-to-end autonomous driving that generates reliable driving trajectories by integrating hierarchical visual input, ego state, and high-level language commands.
    
    \item We develop a multi-stage training strategy that aligns structured 2D and 3D visual features into a unified semantic space, enabling naive VLMs to generate spatially-grounded actions in complex driving scenarios.

    % We propose a multi-stage training strategy with hierarchical vision-language alignment, projecting structured 2D/3D visual tokens into a unified semantic space and extending naive VLMs to 3D spatially-aware action generation in complex driving scenarios.
    % \item We propose a multi-stage training strategy with a hierarchical vision-language alignment module, enabling the projection of structured 2D and 3D visual tokens into a unified semantic space, and extending naive VLMs from static 2D understanding to 3D spatially-aware action generation in complex driving scenarios.
    
    % \item We propose a multi-stage training strategy to extend naive VLMs from 2D image understanding to 3D spatially-aware action generation in complex driving scenarios.

    % \item We introduce hierarchical vision-language feature alignment module, which enables structured 2D and 3D visual tokens projection into a unified semantic embedding space.
    % , enabling effective fusion of perception and language for planning

    \item We introduce implicit agent–environment–ego interaction modeling into autoregressive LLM-based VLA training as an auxiliary task, enabling the model to learn behaviorally grounded and safety-aware driving actions.
    
    % We incorporate structured agent–environment–ego interaction modeling as an auxiliary training task into autoregressive LLM-based VLA models.
    
    % , significantly improving motion forecasting and planning reliability in complex traffic scenario
\end{itemize}
\section{Related Work}

\subsection{End-to-End Autonomous Driving}
Autonomous driving (AD) evolves through two distinct stages. Traditional approaches rely on a modular design, decomposing the system into perception \cite{bevformer}, prediction \cite{pred1}, and planning \cite{planning1} components. While this structure ensures interpretability and allows for independent optimization, they suffer from cascading errors between stages and are not globally optimized for the final planning objective. In contrast, end-to-end autonomous driving frameworks \cite{hu2023_uniad} address this by jointly optimizing perception, prediction, and planning within a unified neural network. These models learn driving policies directly from raw sensor inputs, which improves the model's adaptability to diverse driving conditions. More recent approaches introduce diffusion models \cite{diffusiondrive} and unified scene representations \cite{jia2025drivetransformer} to further enhance the effectiveness and robostness. However, existing end-to-end methods still face semantic reasoning bottlenecks, as they struggle to fully comprehend high-level scene semantics, infer complex agent interactions, and adapt to dynamic task requirements. Moreover, their decision-making processes remain opaque, making it difficult to diagnose failure cases, especially in long-tail or unseen scenarios.

\subsection{Large Vision Language Models} 

Large Language Models demonstrated strong emergent capabilities in in-context learning, instruction following, and reasoning \cite{touvron2023llamaopenefficientfoundation,qwen2.5}. By training on vast amounts of Internet-scale data, these models acquire extensive world knowledge and exhibit strong adaptability across diverse tasks. Their success has also driven the rise of large VLMs, which extend these capabilities into cross-modal reasoning by integrating vision encoders with language models. State-of-the-art VLMs such as GPT-4V \cite{openai2024gpt4technicalreport}, LLaVA \cite{liu2024llavanext}, and Qwen-VL \cite{Qwen-VL} demonstrate strong visual understanding and multimodal reasoning in open-domain tasks. However, these models are primarily trained on static 2D images or videos and exhibit limited spatial reasoning in dynamic 3D driving environments. Moreover, VLMs are prone to hallucinations and generally over-confident but incorrect descriptions, which pose serious risks in safety-critical planning scenarios. Recently, Vision-Language Action models have emerged to directly predict actions from visual inputs, demonstrating strong performance in robotic manipulation tasks \cite{kim24openvla}. Currently, the application of such language-conditioned end-to-end action generation in autonomous driving remains underexplored. Yet, these methods are mostly limited to static setups and lack driving-specific 3D spatial design.

\begin{figure}[t]
    \centering

    \begin{minipage}{0.9\linewidth}
        \centering
        \includegraphics[width=\linewidth]{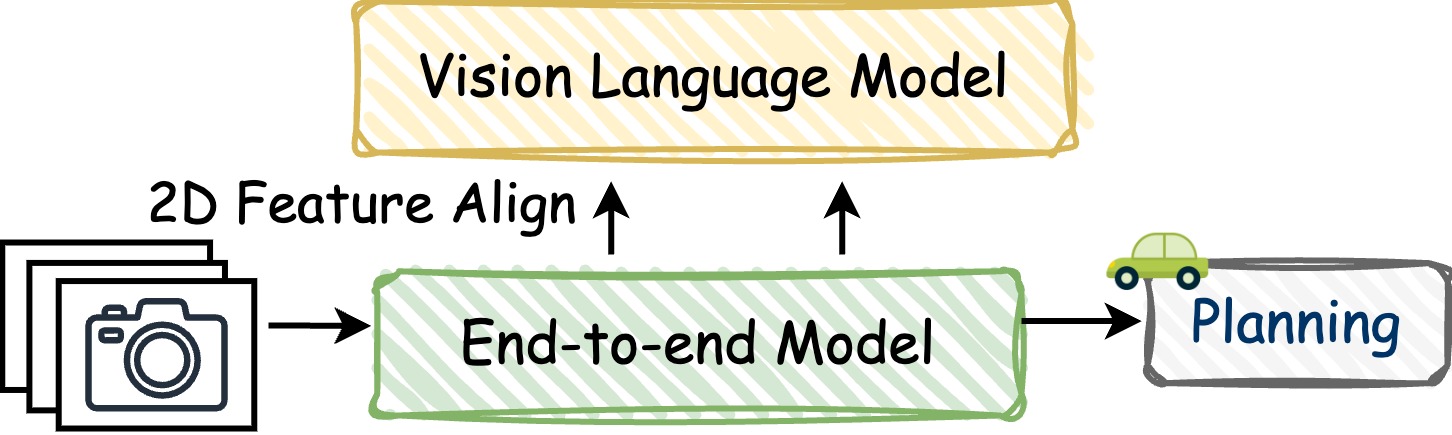}
        \label{fig:related_worka}
        \caption*{(a) VLM as additional Caption or QA Head.}
    \end{minipage}\hfill
    \begin{minipage}{0.9\linewidth}
        \centering
        \includegraphics[width=\linewidth]{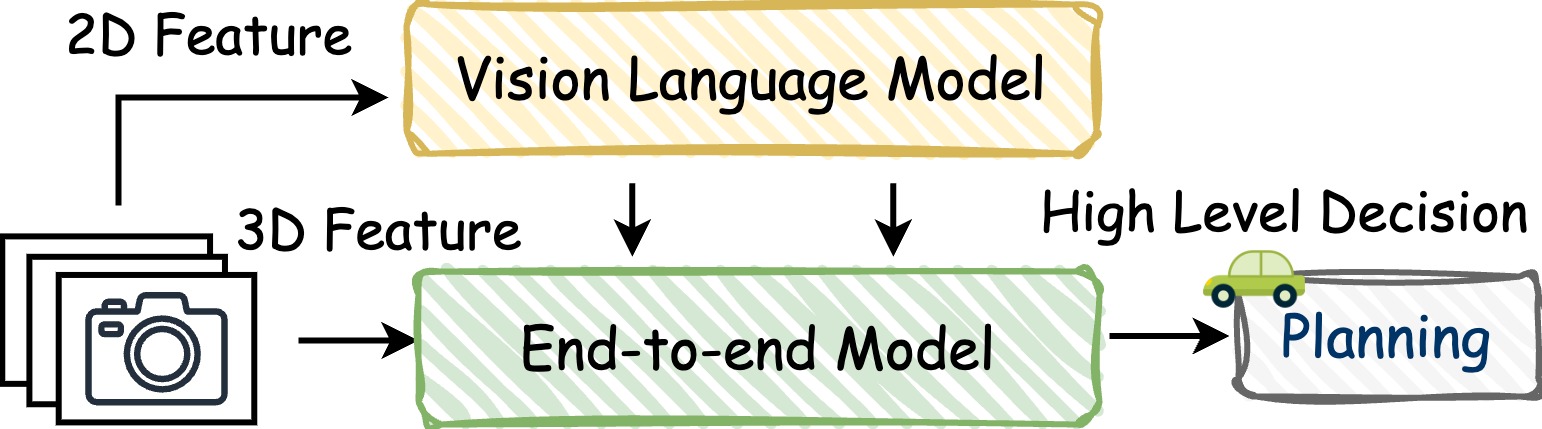}
        \label{fig:related_workb}
        \caption*{(b) VLM as high-level driving decision-maker.}
    \end{minipage}

    \par\medskip

    \begin{minipage}{0.9\linewidth}
        \centering
        \includegraphics[width=\linewidth]{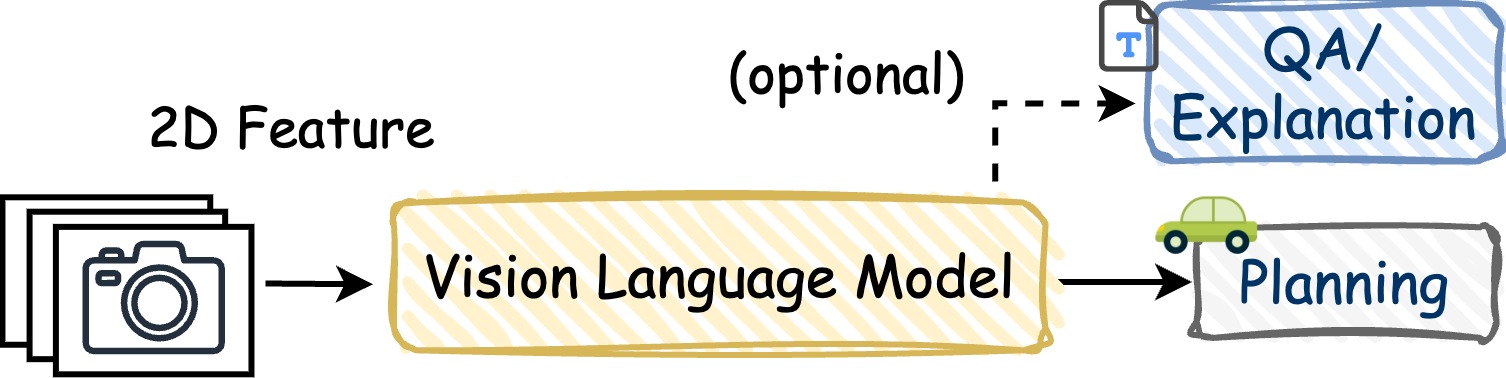}
        \label{fig:related_workc}
        \caption*{(c) Native 2D VLM for end-to-end driving.}
    \end{minipage}\hfill
    \begin{minipage}{0.9\linewidth}
        \centering
        \includegraphics[width=\linewidth]{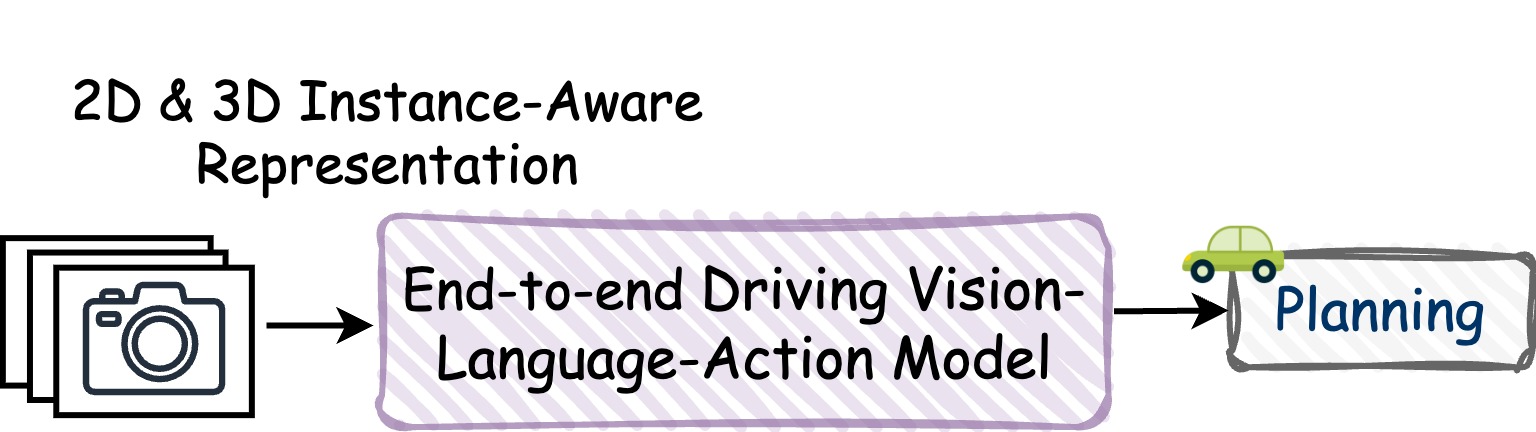}
        \label{fig:related_workd}
        \caption*{(d) 3D spatial-aware driving VLA (ours).}
    \end{minipage}

    \caption{Taxonomy of vision-language model applications in end-to-end autonomous driving.}
    \label{fig:related_work}
\end{figure}

\subsection{Vision Language Models in Autonomous Driving}

% \begin{figure*}[t]
%   \centering
%   \includegraphics[width=0.95\textwidth]{sec_aaai/fig/drivevla-ModelArc.jpg}
%   \caption{OpenDriveVLA leverages open-source pre-trained language foundation models to generate driving actions conditioned on 3D environmental perception, ego vehicle states, and driver commands.}
%   \label{fig:modelarc}
% \end{figure*}
 
VLMs have been applied to various autonomous driving tasks, including perception, scene description, synthetic data generation, and high-level decision-making \cite{vlmadsurvey}. These efforts aim to enhance interpretability, data efficiency, and instruction-following capabilities in driving models. We categorize recent works into 4 paradigms, as illustrated in Figure \ref{fig:related_work}. One line of research in Fig.\ref{fig:related_work} (a) integrates language heads, such as captioning or question-answering modules, into driving models to enhance the interpretability \cite{hintad-nux}. The second category in Fig.\ref{fig:related_work} (b) employs vision language models to generate high-level driving instructions, such as directional commands or abstract maneuvers, which are subsequently interpreted by separate planning modules into low-level controls \cite{jiang2024senna,tian2024drivevlm,wang2023drivemlm}. It's also usually formed as a fast-slow dual system. This design allows VLMs to make independent semantic reasoning, but retains a separate module for end-to-end driving planning, making joint optimization challenging. The third line in Fig.\ref{fig:related_work} (c) applies native VLMs with 2D visual tokens to produce driving actions, and optionally scene captions or QA responses \cite{adapt,drivegpt4}. These methods \cite{leapAD,10655906,fu2025orionholisticendtoendautonomous} process 2D images without explicit modeling of the instance, 3D spatial layout, and inter-agent interactions in the driving scene. It limits their spatial reasoning ability and understanding of agent dynamics in complex traffic environments. Recent studies \cite{favero2024multimodalhallucinationcontrolvisual2} further indicate that such instance-agnostic approaches are more prone to hallucinate, often producing overconfident or semantically inconsistent text. In this work, we investigate how to extend 2D VLMs by explicitly modeling 3D instance-aware and spatial-aware scene representations into an end-to-end autonomous driving framework, as shown in Fig.\ref{fig:related_work}(d). Notably, we focus on fully differentiable end-to-end models in this work, while LLM-based agentic driving systems, such as \cite{wang2024omnidrive,sima2023drivelm}, fall outside the scope of our study.

\section{OpenDriveVLA}
\begin{figure*}[t!]
  \begin{center}
  \includegraphics[width=0.98\textwidth]{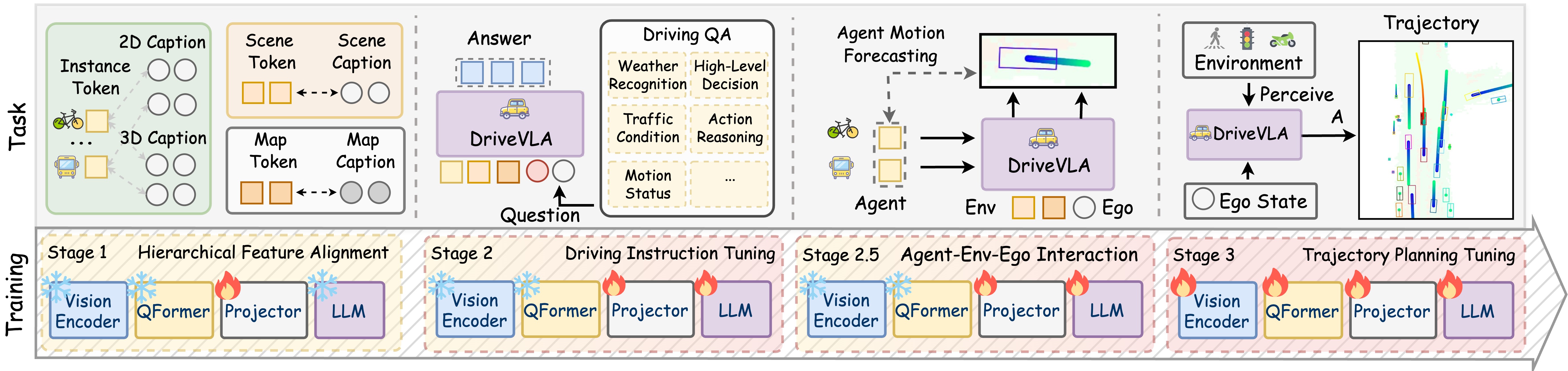}
    \captionof{figure}{Illustration of main training stages on OpenDriveVLA. Stage 1: Hierarchical Feature Alignment. Stage 2: Driving Instruction Tuning. Stage 2.5: Agent-Env-Ego Interaction Modeling. Stage 3: Trajectory Planning Tuning. }
    \label{fig:training_stages}
  \end{center}
\end{figure*}

% \subsection{Overview}
The overall architecture of OpenDriveVLA is shown in Figure \ref{fig:title_figure}, with its multi-stage training process further detailed in Figure~\ref{fig:training_stages}. OpenDriveVLA uses a pre-trained vision encoder to extract tokenized environmental representations from multi-view images. These visual tokens are then aligned into the textual domain through cross-modal learning. After alignment, it undergoes driving instruction tuning, followed by agent-ego-environment interaction modeling. Finally, OpenDriveVLA is trained end-to-end to predict the ego vehicle's future trajectory, guided by the aligned visual-language tokens and driving instructions.

% \begin{figure*}[t!]
%   \begin{center}
%   \includegraphics[width=\textwidth,height=10cm,keepaspectratio]{fig/drivevla-ModelArc.jpg}
%     \captionof{figure}{OpenDriveVLA leverages open-source pre-trained language foundation models to generate driving actions conditioned on 3D environmental perception, ego vehicle states, and driver commands.}
%     \label{fig:title_figure}
%   \end{center}
% \end{figure*}

% It achieves leading performance in both open-loop planning and driving-related question answering, demonstrating its proficiency in driving action planning and scene understanding.
 
% \subsection{Vision Task Pre-training}

\subsection{3D Visual Environmental Perception}
\label{3dvisualenvpercep}
Recent VLM-based autonomous driving methods typically rely on pretrained 2D visual encoders \cite{siglip}, where visual token selection and attention are indirectly guided through language supervision. While effective in open-domain vision-language applications, this design lacks explicit 3D spatial grounding and structured instance-level attention, which can lead to severe hallucinations in safety-critical driving scenarios \cite{drivebench}. To mitigate this, OpenDriveVLA adopts a visual-centric query module, where the model first learns to focus on driving-relevant objects and map tokens through 3D vision tasks, ensuring reliable visual token proposal.

Specifically, given a set of multi-view images ${I} = \{I^i\}_{i=1}^N$, the visual module first extracts multi-scale 2D features from each image using a shared 2D backbone, denoted as $f_{2D}$. These 2D features are then aggregated across views and lifted into BEV space, producing the BEV feature $f_{bev}$. To obtain structured environmental representations, we adopt three visual query modules: Global Scene Sampler $\mathcal{Q}_{\mathrm{scene}}$, Agent QueryTransformer $\mathcal{Q}_{\mathrm{agent}}$, and Map QueryTransformer $\mathcal{Q}_{\mathrm{map}}$. Each module extracts tokens focusing on a specific semantic aspect of the driving environment. Global Scene Sampler encodes the surrounding driving scene context from multi-view 2D features, producing the scene token $v_{scene} = \mathcal{Q}_{\mathrm{scene}}(f_{2D})$. Agent QueryTransformer detects and tracks dynamic agents within the scene, extracting agent-centric tokens $\{v_{agent}^i\}_{i=1}^{N_a} = \mathcal{Q}_{\mathrm{agent}}(f_{bev})$, where $N_a$ denotes the number of detected agents. In parallel, Map QueryTransformer extracts static structural information, such as lane boundaries and drivable areas, forming the map token $v_{map} = \mathcal{Q}_{\mathrm{map}}(f_{bev})$. Through vision-centric perception tasks, including 3D detection, tracking, and segmentation, the visual encoder produces structured environmental tokens that capture both dynamic agent behaviors and static map structures in a spatially grounded manner. The output tokens, denoted as $\textbf{V}_{env} = \{v_{scene}, v_{agent}, v_{map}\}$, serve as visual environment representation of the subsequent stages.

% The projected features $z_{scene}$, $z_{agent}^i$, and $z_{map}$ are embedded into the LLM’s input space, enabling the LLM to process visual tokens directly in its native word embedding space.

\subsection{Stage 1 - Hierarchical Vision-Language Alignment}
\label{hierachvlalignment}
To bridge the modality gap between the extracted visual tokens and the word embedding space of a pre-trained LLM, we adopt a hierarchical vision-language feature alignment strategy. Given the visual tokens extracted from the 3D visual perception module, we introduce three token-specific projectors
$\{\Phi_{\mathrm{scene}}, \Phi_{\mathrm{agent}}, \Phi_{\mathrm{map}}\}$. During training, each active agent query from the 3D detection and tracking task denoted as $v_{agent}^i$, is also matched to its corresponding ground-truth caption $\mathbf{X}_{agent}^i$. These captions provide detailed descriptions, including 2D appearance descriptions and 3D spatial positions. For scene and map tokens, which encode holistic spatial context and static structural properties, a sample-wise alignment is applied, where each token is matched to a scene-level caption $\mathbf{X}_{scene}$ or $\mathbf{X}_{map}$. The scene token $v_{scene}$ captures the global 2D environmental context, while the map token $v_{map}$ encodes structural elements such as lane topology, road boundaries, and drivable areas. Each of these tokens is aligned to its corresponding caption, denoted as $\mathbf{X}_{scene}$ and $\mathbf{X}_{map}$. During this stage, both the visual encoder and LLM remain frozen to preserve pretrained semantics, with only the token-specific projectors being trainable. The forward alignment step is formulated as follows:
\begin{equation}
\hat{\mathbf{X}}_k = \mathrm{LLM}\left(\Phi_k (v_k)\right), \quad k \in \{\mathrm{scene}, \mathrm{map}\}
\end{equation}
\begin{equation}
\hat{\mathbf{X}}_{agent}^i = 
\mathrm{LLM}\left(\Phi_{\mathrm{agent}} (v_{agent}^i)\right), \quad i = 1, \dots, N_a
\end{equation}

\subsection{Stage 2 - Driving Instruction Tuning}
\label{drivinginstructiontuning}

We distill high-level driving knowledge into the model via supervised instruction tuning, enabling it to internalize semantic reasoning patterns during training. This avoids costly chain-of-thought (CoT) reasoning at inference time and balances planning efficacy with runtime efficiency.

During the tuning process, driving knowledge from the language domain is injected into the model using a curated driving instruction QA dataset. The dataset covers a wide range of driving-related reasoning, including perception understanding, motion prediction, attention allocation, action reasoning, and high-level decision-making. By training on this diverse set of driving queries, OpenDriveVLA learns to contextualize the driving scene, follow commands, and generate semantically and behaviorally grounded planning decisions. We formulate the tuning data as instruction-response pairs $\{\mathbf{X}_{input}, \mathbf{X}_{answer}\}$, where $\mathbf{X}_{input} = (\textbf{V}_{env}, \textbf{S}_{ego},\mathbf{X}_{query})$.
Here, $\mathbf{X}_{query}$ denotes the driving-related question, and $\textbf{S}_{ego}$ encodes the textual ego vehicle state. Given this multimodal input, the LLM autoregressively learns to generate the target response. During instruction tuning, the visual encoder remains frozen while the token-specific projectors and the LLM are set to be trainable. The instruction prediction process is as:

\begin{equation}
    \hat{\mathbf{X}}_{answer} = \mathrm{LLM}\left(\textbf{V}_{env}, \textbf{S}_{ego},\mathbf{X}_{query}\right)
\end{equation}

% We introduce a structured driving instruction tuning stage to distill high-level human driving knowledge into DriveVLA and enhance its ability to reason in alignment with human driving behavior. During this process, driving knowledge from the language domain is injected into the model by leveraging a curated driving instruction QA dataset. The questions and answers cover a comprehensive range of driving-related reasoning, including perception understanding, motion prediction, attention allocation, action reasoning, and high-level decision-making. Through this instruction tuning, DriveVLA learns how to understand the driving scene, follow commands, and make contextually appropriate driving decisions.

% $\mathcal{Z}_{env}$ represents the projected environmental visual tokens $\mathcal{Z}_{env} = \{z_{scene}, z_{agent}, z_{map}\}$.

\subsection{Stage 2.5 - Agent Environment Ego Interaction}
\label{agentmotionprediction}
Reliable trajectory planning in autonomous driving necessitates a spatially grounded 3D representation of the environment. Beyond perception, it must also understand dynamic interactions between the ego vehicle and surrounding agents. Effective interaction modeling is essential to ensure that planned trajectories are both feasible and collision-free under real-world driving constraints. However, existing pre-trained LLMs lack an inherent inductive bias for spatial reasoning in 3D driving scenes, as they are predominantly trained on 2D vision-language and text-based datasets. We introduce a conditional agent trajectory forecasting task as an auxiliary objective, encouraging the model to learn spatially grounded interaction priors. During this stage, OpenDriveVLA captures the underlying structure of multi-agent dynamics, enhancing its capability for scene-aware trajectory generation and improving decision-making in complex traffic scenarios.

Given scene and map tokens, as well as the ego vehicle state $\textbf{S}_{ego}$, the LLM predicts the future motion of each detected agent based on its projected visual embedding $\Phi_{\mathrm{agent}} (v_{agent}^i)$. The future motion of agent $a_i$ is represented as a sequence of waypoints $\mathcal{W}^i_{a}$. The predicted trajectory is conditioned on the scene context, map structure, and ego vehicle state, enabling OpenDriveVLA to infer interaction-aware and spatially grounded motion sequences. The learning objective for the $i$-th agent is formulated as:

\begin{equation}
\mathrm{max} \prod_{t=1}^{T} p\left(w^i_t \mid w^i_{1:t-1}, \textbf{V}_{env}, \textbf{S}_{ego}, \Phi_{\mathrm{agent}} (v_{agent}^i) \right)
% \hat{\mathcal{T}}_{traj} = \operatorname*{argmax} p\left(\mathcal{T}_{traj} \mid \textbf{V}_{env}, \textbf{S}, \textbf{X}_{dri}\right)
\end{equation}

This provides OpenDriveVLA with essential spatial priors, enabling it to bridge the gap between high-level semantic reasoning and physically grounded motion planning.

% \begin{equation}
% \hat{\mathcal{W}}_i = \operatorname*{argmax}_{\mathcal{W}_i} \prod_{t=1}^{T} P(w_i^t \mid w_i^{1:t-1}, v_{scene}, v_{map}, q_{ego})
% \end{equation}

% We introduce a learnable special token $<\text{trajectory}>$ to represent the ego vehicles within the shared text embedding space. This token acts as the vehicle query to interact with other agents, maps, and environments. 

\begin{table*}[t]
    \centering
    \resizebox{\linewidth}{!}{ 
    \renewcommand{\arraystretch}{1.2}
    \begin{tabular}{lcccccccc|cccccccc|cc}
    \toprule
     \multirow{3}{*}{\textbf{Method}} & \multicolumn{8}{c}{\textbf{ST-P3 metrics}} & \multicolumn{8}{c}{\textbf{UniAD metrics}} & \multirow{3}{*}{{\textbf{LLM}}} & \multirow{3}{*}{{\textbf{Input}}} \\
    \cmidrule(lr){2-9} \cmidrule(lr){10-17}
    & \multicolumn{4}{c}{\textbf{L2 (m) ↓}} & \multicolumn{4}{c}{\textbf{Collision (\%) ↓}} & \multicolumn{4}{c}{\textbf{L2 (m) ↓}} & \multicolumn{4}{c}{\textbf{Collision (\%) ↓}} &  & \multirow{2}{*}{\textbf{}} \\
    \cmidrule(lr){2-5} \cmidrule(lr){6-9} \cmidrule(lr){10-13} \cmidrule(lr){14-17}
    & 1s & 2s & 3s & Avg. & 1s & 2s & 3s & Avg. & 1s & 2s & 3s & Avg. & 1s & 2s & 3s & Avg. & & \\
    \midrule
    \multicolumn{19}{c}{\textbf{None-Autoregressive Methods}} \\
    \midrule
    ST-P3 \cite{stp3} & 1.33 & 2.11 & 2.90 &   2.11 & 0.23 & 0.62 & 1.27 &   0.71 & - & - & - &   - & - & - & - &   - & - & Visual \\ 
    % 628.3
    
    VAD \cite{vad} & 0.17 & 0.34 & 0.60 &   0.37 & 0.07 & 0.10 & 0.24 &   0.14 & - & - & - &   - & - & - & - &   - & - & Visual \\
    % 224.3
    Ego-MLP \cite{ego-mlp} & 0.46 & 0.76 & 1.12 &   0.78 & 0.21 & 0.35 & 0.58 &   0.38 & - & - & - &   - & - & - & - &   - & - & Ego \\
    
    UniAD \cite{hu2023_uniad} & 0.44 & 0.67 & 0.96 & 0.69 & 0.04 & 0.08 & 0.23 & 0.12 & 0.48 & 0.96 & 1.65 &   1.03 & 0.05 & 0.17 & 0.71 &   0.31 & - & Visual \\
    
    InsightDrive \cite{song2025insightdriveinsightscenerepresentation} & 0.23 & 0.41 & 0.68 & 0.44 & 0.09 & 0.10 & 0.27 & 0.15 & 0.30 & 0.72 & 1.41 &  0.81 & 0.08 & 0.15 & 0.84 &  0.36 & - & Visual \\
    
    % BEV-Planner \cite{bevplanner} & 0.30 & 0.53 & 0.83 &   0.55 & 0.10 & 0.37 & 1.30 &   0.59 & - & - & - &   - & - & - & - &   - & - & - \\
    FF \cite{planning1} & - & - & - &   - & - & - & - &   - & 0.55 & 1.20 & 2.54 &   1.43 & 0.06 & 0.17 & 1.07 &   0.43  & - & LiDAR \\
    
    % 555.6
    EO \cite{EO} & - & - & - &   - & - & - & - &   - & 0.67 & 1.36 & 2.78 &   1.60 & 0.04 & \textbf{0.09} & 0.88 &   0.33 & - & LiDAR \\
    
    \midrule
    \multicolumn{19}{c}{\textbf{Autoregressive Methods}} \\
    \midrule
    % Atlas \cite{atlas} & 0.52 & 0.97 & 1.53 &   1.00 & 0.15 & 0.31 & 0.70 &   0.38 & - & - & - &   - & - & - & - &   - & - & - \\
    
    GPVL \cite{GPVL} & 0.21 & 0.39 & 0.69 &   0.43 & 0.07 & 0.09 & 0.27 &   0.14 & - & - & - &   - & - & - & - &   - & BERT & Textual \\
    
    DriveVLM \cite{tian2024drivevlm} & 0.18 & 0.34 & 0.68 &   0.40 & 0.10 & 0.22 & 0.45 &   0.27 & - & - & - &   - & - & - & - &   - & Qwen-VL-7B & Visual \\
        
    GPT-Driver \cite{mao2024gptdriver} & 0.20 & 0.40 & 0.70 &   0.44 & 0.04 & 0.12 & 0.36 &   0.17 & 0.27 & 0.74 & 1.52 &   0.84 & 0.07 & 0.15 & 1.10 &   0.44 & GPT-3.5 & Textual \\
    
    RDA-Driver \cite{RDA-Driver} & 0.17 & 0.37 & 0.69 &   0.40 & 0.01& \textbf{0.05} & 0.26 &   0.10 & 0.23 & 0.73 & 1.54 &   0.80 & \textbf{0.00} & 0.13 & 0.83 &   0.32 & LLaVa-7B & Visual \\
    
    OminiDrive \cite{wang2024omnidrive} & 0.14 & 0.29 & 0.55 & 0.33 & \textbf{0.00} & 0.13 & 0.78 & 0.30 & -  & - & - &  - & - & - & - &   - &  LLaVA-7B & Visual \\
    
    EMMA \cite{hwang2024emmaendtoendmultimodalmodel} & \textbf{0.14} & \textbf{0.29} & \textbf{0.54} & \textbf{0.32} & - & - & - &   - & - & - & - & - & - & - & - & - & \textbf{Gemini} & Visual \\
    
    % EMMA \cite{hwang2024emmaendtoendmultimodalmodel} & \textbf{0.14} & \textbf{0.29} & \textbf{0.54} & \textbf{0.32} & - & - & - &   - & - & - & - & - & - & - & - & - & \textbf{Gemini} \cite{geminiteam2024gemini15unlockingmultimodal} & Visual \\

    OpenEMMA \cite{xing2025openemmaopensourcemultimodalmodel} & 1.45 & 3.21 & 3.76 & 2.81 & - & - & - &   - & - & - & - &   -  & - & - & - & - & Qwen-VL-7B & Visual \\
    
    DME-Driver \cite{dmedriver} & - & - & - &   - & - & - & - &   - & 0.45 & 0.91 & 1.58 &   0.98 & 0.05 & 0.28 & \textbf{0.55} &  0.29 & LLaVa-7B & Visual \\

    \midrule
 
    OpenDriveVLA-0.5B (Ours) & 0.15 & 0.32 & 0.57 &   0.35 & 0.01& 0.06 & \textbf{0.20} &   \textbf{0.09} & 0.21 & 0.60 & 1.22 &   0.68 & \textbf{0.00} & 0.15 & 0.63 &    0.26 & Qwen2.5-0.5B & Visual \\
 
    OpenDriveVLA-3B (Ours) & \textbf{0.14} & 0.30 & 0.55 &   0.33 & 0.02 & 0.07 & 0.22 &  0.10 & \textbf{0.19} & \textbf{0.58} & 1.24 &   0.67 & 0.02 & 0.18 & 0.70 &    0.30 & Qwen2.5-3B & Visual \\
 
    OpenDriveVLA-7B  (Ours) & 0.15 & 0.31 & 0.55 &    0.33 & 0.01& 0.08 & 0.21 &   0.10 & 0.20 & \textbf{0.58} & \textbf{1.21} &   \textbf{0.66} & \textbf{0.00} & 0.22 & \textbf{0.55} & \textbf{0.25}  & Qwen2.5-7B & Visual \\
    \bottomrule
    \end{tabular}}
    \caption{Open-Loop planning performance comparison of different driving models, including both autoregressive methods and non-autoregressive methods. OpenDriveVLA shows powerful planning ability and achieves best-in-class results among open-source models, even with the 0.5B version. We refer to the result summary from \cite{song2025insightdriveinsightscenerepresentation,mao2024gptdriver,GPVL,RDA-Driver}.}
    \label{tab:open-loop-planning}
\end{table*}
% [x] introduce coordinate..

\subsection{Stage 3 - End-to-end Trajectory Planning Tuning}
\label{drivingtrajpred}
In this stage, OpenDriveVLA predicts ego trajectories as discrete waypoint sequences within a short horizon, denoted as $\mathcal{W}_{ego} = \{w_1, w_2, \dots, w_T\}$. Each waypoint $w_t$ represents the 2D coordinates $(x_t, y_t)$ of the ego vehicle at time step $t$. The waypoints are tokenized into a sequence of discrete textual tokens for autoregressive generation in the LLM: $\mathcal{T}{traj} = \mathrm{Tokenizer}(\mathcal{W}_{ego})$. The generation process is then cast as a causal sequence prediction task, where each token is predicted in a causal manner, conditioned on the visual perception tokens $\textbf{V}_{env}$, the ego state $\textbf{S}_{ego}$, and the driving command $\textbf{X}_{dri}$. 

% \begin{equation}
% \hat{\mathcal{T}}_{traj} = \textit{argmax}_{\textbf{T}_{traj}} \prod_{t=1}^{T} p\left(w_t \mid w_{1:t-1}, \textbf{V}_{env}, \textbf{S}_{ego}, \textbf{X}_{dri}\right)
% \end{equation}

\begin{equation}
\resizebox{\linewidth}{!}{$
\hat{\mathcal{T}}_{traj}
= \textit{argmax}_{\textbf{T}_{traj}}
\prod_{t=1}^{T}
p\left(
w_t \mid
w_{1:t-1},
\textbf{V}_{env},
\textbf{S}_{ego},
\textbf{X}_{dri}
\right)
$}
\end{equation}

% \begin{equation}
% \begin{aligned}
% \hat{\mathcal{T}}_{traj}
% &= \textrm{argmax}_{\textbf{T}_{traj}}
% \prod_{t=1}^{T}
% p\big(
% w_t \mid w_{1:t-1}, \\
% &\quad \textbf{V}_{env}, \textbf{S}_{ego}, \textbf{X}_{dri}
% \big)
% \end{aligned}
% \end{equation}

The  entire pipeline, including the 3D visual encoder, cross-modality projectors, and LLM, is jointly optimized end-to-end during training, with the 2D encoder kept frozen. At inference, the model autoregressively generates the tokenized trajectory $\hat{\mathcal{T}}_{traj}$, which is then decoded back into numerical waypoints:

\begin{equation}
\hat{\mathcal{W}}_{ego} = \mathrm{Decoder}(\hat{\mathcal{T}}_{traj})
\end{equation}

% [] with agent matching + with filtered object..

% [] matching strategy.. explain how to match.. 

\section{Experiments}

\subsection{Training Datasets}
We curate the training data of OpenDriveVLA based on its distinct training phases, drawing from: TOD3Cap \cite{TOD3Cap}, nuCaption \cite{nucaption}, nuScenesQA \cite{nuscenesqa}, nuX \cite{hintad-nux}, and GPT-Driver \cite{mao2024gptdriver}. We conduct experiments on nuScenes \cite{nuScenes}, following standard data split into training and validation sets. OpenDriveVLA is trained using the training set paired with corresponding QA captions, while the validation set is exclusively used for performance evaluation to ensure fair comparisons with prior works. The details of training data can be found in supplementary materials.

% DriveVLA adopts a 3D perception visual encoder, Agents QueryTransformer, and Map QueryTransformer which are pre-trained through vision-centric multi-tasks, including 3D detection, tracking, and map segmentation following \cite{hu2023_uniad}. 
\noindent\textbf{Hierarchical Vision-Language Alignment.} For agent-level caption, we post-process data from \cite{TOD3Cap}, which provides the 2D visual description of individual objects. To further enhance spatial grounding, each object caption is augmented with its corresponding BEV coordinates, enabling the model to associate object attributes with precise spatial locations. For scene tokens, we process multi-view scene descriptions from \cite{nucaption}, merging them into unified summaries that describe the driving environment across all camera views. For map tokens, structured language descriptions are derived from ground-truth annotations, translating map elements such as lane dividers, crosswalks, and road boundaries into descriptive text. 

% The generated map caption captures both semantic categories and quantitative spatial attributes to enrich the language-aware map representation.

\noindent\textbf{Driving Instruction Tuning.} We adopt multiple instruction-oriented datasets derived from nuScenes to inject driving-specific knowledge into OpenDriveVLA. We unify several datasets into a standardized instruction-based QA format, including driving-related question-answer pairs collected from nuCaption \cite{nucaption}, nuScenesQA \cite{nuscenesqa}, and nuX \cite{hintad-nux} dataset. Each QA pair is conditioned on structured environmental visual tokens and the ego vehicle state, ensuring consistency across different data sources. This multimodal instruction tuning process allows OpenDriveVLA to effectively ground language understanding into both environmental perception and scene understanding, bridging perception, reasoning, and action within the language space.
% , enabling the implicit driving knowledge installation and user vehicle interaction
 % optimize the model's ability to perceive, predict, focus, and plan on the textual space of LLM.

\noindent\textbf{Motion Forecasting and Trajectory Prediction.} We formulate both agent motion forecasting and ego trajectory planning in the ego system, where the model directly predicts future displacements within each entity’s local coordinate frame relative to the ego vehicle for planning and relative to each agent for forecasting. This formulation captures motion dynamics in a spatially consistent manner across all entities. Following \cite{mao2024gptdriver}, the ego vehicle state is encoded as textual input to ensure ego awareness throughout the training process. Both tasks predict 3-second future trajectories, sampled at 0.5-second intervals, resulting in 6 waypoints per trajectory.

\begin{table*}[t!]
\centering
\resizebox{0.98\linewidth}{!}{%
\begin{tabular}{l|ccccc|cccccccc}
\toprule
\multirow{2}{*}{\textbf{Method}} & \multicolumn{5}{c|}{\textbf{nu-Caption}} & \multicolumn{8}{c}{\textbf{nuScenes-QA}} \\
\cmidrule(lr){2-6} \cmidrule(lr){7-14}
& \textbf{BL-1} & \textbf{BL-2} & \textbf{BL-3} & \textbf{BL-4} & \textbf{BERT-S} 
& \textbf{Ext} & \textbf{Cnt} & \textbf{Obj} & \textbf{Sts} & \textbf{Cmp} & \textbf{H0} & \textbf{H1} & \textbf{Acc} \\
\midrule
Mini-GPT4 \cite{zhu2024minigpt4} & 15.0 & 6.8 & 3.7 & 2.6 & 84.4 & - & - & - & - & - & - & - & - \\
Instruct-BLIP \cite{dai2023instructblip} & 18.7 & 13.4 & 7.4 & 5.2 & 85.9 & - & - & - & - & - & - & - & - \\
LLaMA-AdapV2 \cite{gao2023llama-adapter-v2} & 30.2 & 17.3 & 10.4 & 7.5 & 86.5 & 19.3 & 2.7 & 7.6 & 10.8 & 1.6 & 15.1 & 4.8 & 9.6 \\
LLaVA1.5 \cite{llava1_5} & 20.0 & 12.1 & 8.6 & 5.4 & 85.0 & 45.8 & 7.7 & 7.8 & 9.0 & 52.1 & 25.7 & 41.5 & 26.2 \\
LiDAR-LLM \cite{nucaption} & 41.0 & 30.0 & 23.4 & 19.3 & 91.3 & 74.5 & 15.0 & 37.8 & 45.9 & 57.8 & - & - & 48.6 \\
BEVDet+BUTD \cite{nuscenesqa} & - & - & - & - & - & 83.7 & 20.9 & 48.8 & 52.0 & 67.7 & - & - & 57.0 \\
\midrule
OpenDriveVLA-0.5B (Ours) & 47.2 & 35.8 & 29.4 & 25.2 & 91.9 & 83.9 & 22.0 & 50.2 & \textbf{57.0} & 68.4 & 62.3 & \textbf{56.5} & 58.4 \\
% \midrule
OpenDriveVLA-3B (Ours) & 48.3 & 36.9 & 30.3 & 26.1 & 92.0 & 84.0 & 22.3 & \textbf{50.3} & 56.9 & 68.5 & \textbf{62.6} & \textbf{56.5} & \textbf{58.5}  \\
% \midrule
OpenDriveVLA-7B (Ours) & \textbf{49.6} & \textbf{38.3} & \textbf{31.9} & \textbf{27.6} & \textbf{92.2} & \textbf{84.2} & \textbf{22.7} & 49.6 & 54.5 & \textbf{68.8} & 62.4 & 56.1 & 58.2 \\
\bottomrule
\end{tabular}%
}
\caption{Performance on nu-Caption \cite{nucaption} and nuScenes-QA \cite{nuscenesqa}. BL-1/2/3/4: BLEU scores. QA metrics report accuracy on five question types: Existence, Counting, Object, Status, and Comparison.}
\label{tab:nuqa_nucap}
\end{table*}

\subsection{Evaluations}
We evaluate OpenDriveVLA on the open-loop planning task of nuScenes benchmark, where the model is reported under both ST-P3 \cite{stp3} and UniAD \cite{hu2023_uniad} settings. The evaluation metrics include L2 displacement errors at 1, 2, and 3 seconds, along with the average collision rate over the prediction horizon. To further assess the scene understanding ability of OpenDriveVLA, we report its QA prediction performance on three driving visual question answering (VQA) datasets directly after the driving instruction tuning stage, i.e., \cite{nucaption}, nuScenesQA \cite{nuscenesqa}, and nuX \cite{hintad-nux}. The VQA evaluation results adopt standard NLG metrics, including BLEU, METEOR, CIDEr, BERT-Score, etc.

 % and a BEV encoder to transform multi-view image features into a structured BEV feature map
\subsection{Implementation Details}
The 3D visual perception module in OpenDriveVLA follows the vision-centric design from \cite{hu2023_uniad}, using a ResNet-101 backbone for 2D feature extraction. The perception backbone is pre-trained via multi-task learning on 3D object detection, object tracking, and map segmentation. The resulting BEV feature map has a spatial resolution of $200 \times 200$. To construct a unified scene representation, the global SceneSampler applies 2D adaptive pooling to each camera view, subsequently concatenating the pooled multi-view features into a global scene token. Agent and map tokens are extracted from the final layer of their respective QueryTransformer modules. Each token type is then mapped into the language space using a separate two-layer MLP with GeLU activation. We adopt Qwen 2.5-Instruct \cite{qwen2.5} as the pre-trained LLM, which undergoes full parameter tuning during training. Training is performed on 4 NVIDIA H100 GPUs with a batch size of 1, completed in approximately two days.  We freeze the 2D backbone during stage 3. During inference, we set the decoding temperature to 0 to ensure deterministic trajectory generation. See supplementary material for detailed training configurations.

\subsection{Main Results}

\noindent\textbf{Open Loop Trajectory Planning.} We evaluate OpenDriveVLA on the open-loop trajectory planning task using both ST-P3 and UniAD metrics, ensuring comprehensive performance assessment across spatial accuracy and collision avoidance. As shown in Table~\ref{tab:open-loop-planning}, OpenDriveVLA achieves state-of-the-art performance across both settings. Specifically, both 3B and 7B version models achieve an average L2 error of 0.33m under ST-P3 metrics, outperforming prior autoregressive language models \cite{mao2024gptdriver,tian2024drivevlm}. On the UniAD metrics, OpenDriveVLA-7B also achieves great performance with an average L2 error of 0.66m. Notably, despite significantly fewer parameters, the 0.5B version still outperforms prior models obviously. 
% , making it an effective and scalable solution for language-guided planning.

% \vspace{1em}

\begin{table}[h!]
\centering
\caption{Performance comparison of OpenDriveVLA on the Nu-X dataset \cite{hintad-nux}.}
\resizebox{0.98\linewidth}{!}{%
\begin{tabular}{l|cccc}
\toprule
\textbf{Models} & \textbf{CIDER} & \textbf{BL-4} & \textbf{METEOR} & \textbf{ROUGE-L} \\
\midrule
Hint-UniAD \cite{hintad-nux} & 21.7 & 4.2 & 12.7 & 27.0 \\
Hint-VAD \cite{hintad-nux} & 22.4 & 4.2 & \textbf{13.2} & 27.6 \\
GPT-4o \cite{drivegpt4} & 19.0 & 4.0 & 10.3 & 24.9 \\
Gemini 1.5 \cite{geminiteam2024gemini15unlockingmultimodal} & 17.6 & 3.4 & 9.3 & 23.4 \\
Vote2CapDETR \cite{vote2cap} & 15.3 & 2.6 & 10.9 & 24.2 \\
TOD\textsuperscript{3}Cap \cite{TOD3Cap} & 14.5 & 2.5 & 10.5 & 23.5 \\
\midrule
\multicolumn{5}{c}{\textbf{OpenDriveVLA}} \\
\midrule
0.5B (Ours) & \textbf{32.3} & \textbf{5.4} & 12.5 & \textbf{27.9} \\
3B (Ours) & 25.5 & 4.3 & 12.8 & 27.8 \\
7B (Ours) & 26.2 & 4.5 & 12.8 & 27.4 \\
\bottomrule
\end{tabular}%
}
\label{tab:nux}
\end{table}

\noindent\textbf{Driving Question Answering.} We access OpenDriveVLA on the driving VQA task across three nuScenes-based datasets (Table~\ref{tab:nuqa_nucap}, Table~\ref{tab:nux}), reporting results after the second stage of training. OpenDriveVLA reaches best-in-class performance across all three datasets, consistently outperforming previous language-enhanced driving models and general-purpose multimodal baselines among most metrics. On nuCaption dataset, it achieves the best captioning performance among all evaluated models, outperforming both general VLMs LLaVA1.5 \cite{llava1_5} and Mini-GPT4 \cite{zhu2024minigpt4}, as well as autonomous driving-specific models such as LiDAR-LLM \cite{nucaption}. For nuScenesQA dataset, OpenDriveVLA also achieves strong performance. Compared to models that directly fuse BEV features with language models such as BEVDet+BUTD \cite{nuscenesqa}, it demonstrates clear advantages in object and status-related questions, which highlights the benefit of its spatially grounded visual-language alignment. Notably, the 0.5B version outperforms even the larger 7B on the Nu-X dataset, which shows its powerful scene-understanding ability even with lightweight LLMs.

\begin{figure*}[t!]
    \centering
    % First row
    % \begin{subfigure}[t]{\linewidth}
    \centering
    \includegraphics[width=0.98\textwidth]{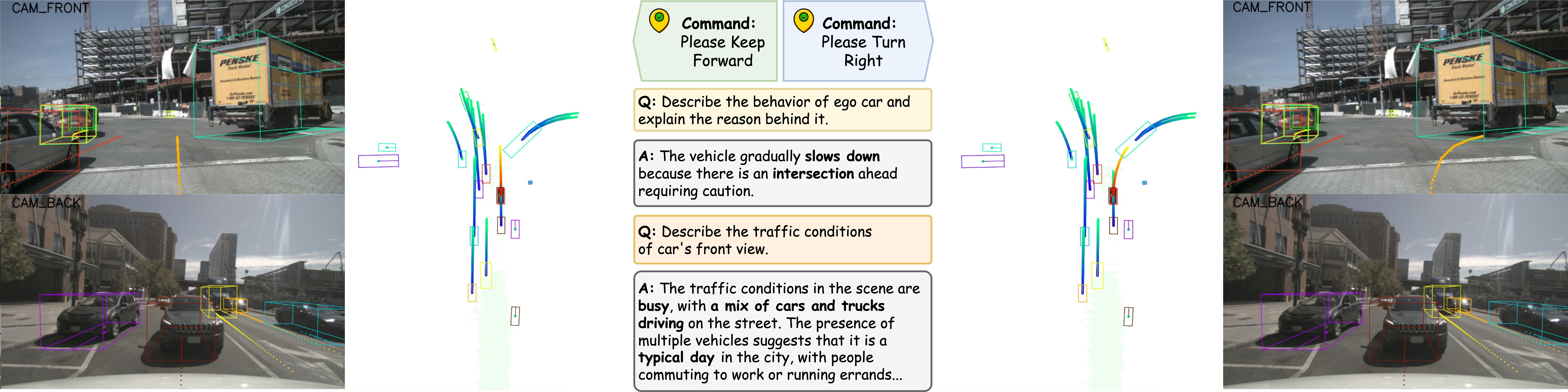}
    % \caption{}
    \label{fig:uniad}
    % \end{subfigure}
    \hfill

    % \vspace{0.1cm}

\caption{Visualization of OpenDriveVLA-7B planning actions under original dataset instruction to keep forward (left) and modified instruction to turn right (right). The QA prediction showcases (middle) are from results reported in Table \ref{tab:nuqa_nucap} and Table \ref{tab:nux}. The agent motion prediction results are visualized after the agent-env-ego interaction stage.}
\label{fig:qualitative1}
\end{figure*}

% \subsection{Ablation Study}
% Our ablation study examines the influence of input modalities on OpenDriveVLA’s trajectory planning. The results in Table \ref{ablative} show that visual inputs enhance the action-making process, while textual high-level commands and historical state information further refine trajectory generation, indicating the contribution of semantic intent and temporal context. Besides, we find that ego-state information plays a significant role in trajectory prediction on nuScenes open-planning benchmark, aligning with prior findings \cite{bevplanner}. Due to the imbalanced distribution of the dataset, where many scenarios involve maintaining the current state, the model tends to over-rely on ego-state history, leading to a bias toward conservative predictions. Hence, to further evaluate the generalization and instructed action-following capability of OpenDriveVLA, we conduct qualitative studies to examine its ability to execute diverse actions in response to different human commands.

\subsection{Ablation Study}
We conduct ablation studies to evaluate the impact of input modalities and our multi-stage training strategy on OpenDriveVLA’s performance. Additionally, we qualitatively assess the model’s ability to follow diverse driving commands.
% We conduct two ablation studies of OpenDriveVLA in terms of input modalities, multi-stage training, and also include a qualitative experiment to show its command-following capability.

\begin{table}[tbhp]
\centering
\resizebox{0.98\linewidth}{!}{
\begin{tabular}{c|c|c|c|c|c|c|c}
\toprule
\multirow{2}{*}{\textbf{Visu}} & \multirow{2}{*}{\textbf{Ego}} & \multirow{2}{*}{\textbf{Hist}} & \multirow{2}{*}{\textbf{Cmd}} & \multicolumn{2}{c|}{\textbf{Avg. Collision (\%) ↓}} & \multicolumn{2}{c}{\textbf{Avg. L2 (m) ↓}} \\
& & & & \textbf{UniAD} & \textbf{ST-P3} & \textbf{UniAD} & \textbf{ST-P3} \\
\midrule
\checkmark & & \checkmark & \checkmark & 0.77 & 0.24 & 1.34 & 0.75 \\
\checkmark & \checkmark &  & \checkmark & 1.14 & 0.49 & 1.30 & 0.75 \\
& \checkmark & \checkmark & \checkmark & 0.29 & 0.10 & 0.77 & 0.39 \\
\checkmark & \checkmark & \checkmark &  & 0.33 & 0.13 & 0.80 & 0.40 \\
\checkmark & \checkmark & \checkmark & \checkmark & 0.26 & 0.09 & 0.68 & 0.35 \\
\bottomrule
\end{tabular}
}
\caption{Ablation study on the effect of different input combinations on OpenDriveVLA-0.5B.}
\label{ablative}
\end{table}

\noindent\textbf{Effect of Input Modalities.} We investigate how individual input components contribute to trajectory planning. Table~\ref{ablative} presents the results of ablating visual perception, ego state, historical trajectory, and high-level language commands. The inclusion of visual inputs significantly boosts overall performance. Adding textual commands and historical information further improves the predictions, emphasizing the value of semantic intent and temporal context. Notably, ego-state features play a critical role in nuScenes open-loop benchmark, consistent with prior findings~\cite{bevplanner}.

% : vision-language alignment, instruction tuning, interaction modeling, and trajectory planning. 

\noindent\textbf{Effect of Multi-Stage Training Strategy.} We evaluate the contribution of each training phase in our staged pipeline incrementally. As shown in Table~\ref{tab:stage_ablation}, each additional stage consistently improves performance, with the most notable reductions in collision rate observed after Hierarchical Vision-Language Alignment and Agent-Environment-Ego Interaction Modeling. These improvements highlight the effectiveness of cross-modal grounding and interaction-aware reasoning in enhancing safety-critical planning behavior.

\begin{table}[tbhp]
\centering
\resizebox{0.98\linewidth}{!}{
\begin{tabular}{c|c|c|c|c|c|c|c}
\toprule
\multicolumn{4}{c|}{\textbf{Training Stage}} & \multicolumn{ 2 }{c|}{\textbf{Avg. Collision (\%) ↓}} & \multicolumn{2}{c}{\textbf{Avg. L2 (m) ↓}} \\
\textbf{ 1 } & \textbf{ 2 } & \textbf{2.5} & \textbf{3} & \textbf{UniAD} & \textbf{ST-P3} & \textbf{UniAD} & \textbf{ST-P3} \\
\midrule
& & & \checkmark & 0.37 & 0.13 & 0.70 & 0.36 \\
\checkmark & & & \checkmark & 0.32 & 0.12 & 0.69 & 0.35 \\
\checkmark & \checkmark & & \checkmark & 0.31 & 0.11 & 0.68 & 0.35 \\
\checkmark & \checkmark & \checkmark & \checkmark & 0.26 & 0.09 & 0.68 & 0.35 \\
\bottomrule
\end{tabular}
}
\caption{Ablation study on the effect of multi-stage training of 0.5B model. Stage 1, 2, 2.5, and 3 correspond to hierarchical feature alignment, driving instruction tuning, Agent-Env-Ego modeling, and trajectory tuning, respectively.}
\label{tab:stage_ablation}
\end{table}

\noindent\textbf{Effect of Driving Command.} Figure \ref{fig:qualitative1} presents the qualitative comparison at an intersection under two different driver instructions: keep forward and turn right, with the right turn as the ground truth. OpenDriveVLA accurately adapts its plan to the given command while maintaining context-aware and environment-consistent behavior, demonstrating robust command-following and generalization in complex scenes. In addition, we visualize the QA predictions for the same scene, showcasing the model’s ability to reason over decision-making and traffic scene understanding.

\section{Conclusion}

In this work, we present OpenDriveVLA, a scalable vision-language action model designed for end-to-end autonomous driving. Built upon pre-trained large language models, OpenDriveVLA generates 3D spatially grounded and semantically consistent driving actions from multimodal inputs. We introduce a hierarchical vision-language feature alignment module and realize agent-env-ego interaction in LLM to enable fine-grained spatial reasoning and dynamic scene understanding. Through multi-stage training paradigm, OpenDriveVLA achieves state-of-the-art performance in open-loop planning and driving-related question answering. Extensive evaluations on nuScenes dataset show its superior trajectory planning capability compared to existing approaches. Our work demonstrates the feasibility of a scalable vision-language-driven approach for autonomous driving and highlights the potential of large language models as a foundation for end-to-end driving action systems.

\clearpage



% \section*{ACKNOWLEDGMENT}

% This research was supported by the Federal Ministry of Education and Research in Germany (BMBF) within the project ”AUTOtech.agil” (Grant Number 01IS22088U). 

%%%%%%%%%%%%%%%%%%%%%%%%%%%%%%%%%%%%%%%%%%%%%%%%%%%%%%%%%%%%%%%%%%%%%%%%%%%%%%%%

% ======== main bibliography (ends main content) ========
{
    \bibliographystyle{IEEEtran}
    \bibliography{main}
}

% ======== Reset TOC for appendix ========
\clearpage
\begingroup
\onecolumn   % ← 保证附录是单栏
\setlength{\cftbeforesecskip}{10pt}
\renewcommand{\cftsecfont}{\normalsize}
\renewcommand{\cftsecpagefont}{\normalsize}
\renewcommand{\numberline}[1]{\hb@xt@3.5em{#1\hfil}\hspace{0.6em}}
\setlength{\cftsecnumwidth}{3.5em}
\setlength{\cftsubsecnumwidth}{4em}
\setcounter{tocdepth}{2}

\begin{center}
{\LARGE Appendix-OpenDriveVLA: Towards End-to-End Autonomous}\\[8pt]
{\LARGE Driving with Large Vision Language Action Model}
\end{center}

\vspace{1.5em}

% ======== 附录正文放在目录之前执行，这样 .toc 会写入 ========
\section{Implementation Details}

% \section{Implementation Details}

\subsection{Model Details}
\subsubsection{Vision Encoder}

To obtain accurate instance-level token representations, one option is to adopt language-guided visual grounding tasks \cite{xiao2024visualgroundingsurvey,visualgrounding1}, where visual regions are aligned with textual descriptions with cross-modality supervision. However, such supervision is often ambiguous and imprecise, especially in complex traffic environments where spatial accuracy is essential. This ambiguity arises from the inherent subjectivity of textual annotations and the weak spatial constraints in general vision-language datasets \cite{zhou2025tumtraf}. Moreover, they typically lack consistent object definitions and fail to capture structured scene semantics, making them suboptimal for autonomous driving applications that demand precise object localization and spatial understanding.

Motivated by these, we adopt a vision-centric pretraining visual encoder and instance token decoder based on high-quality autonomous driving datasets. The overall architecture of the vision encoder and token extraction modules is illustrated in Figure~\ref{fig:stage0_vis1}. 

\begin{figure}[h]
    \centering
    \includegraphics[width=0.9\linewidth]{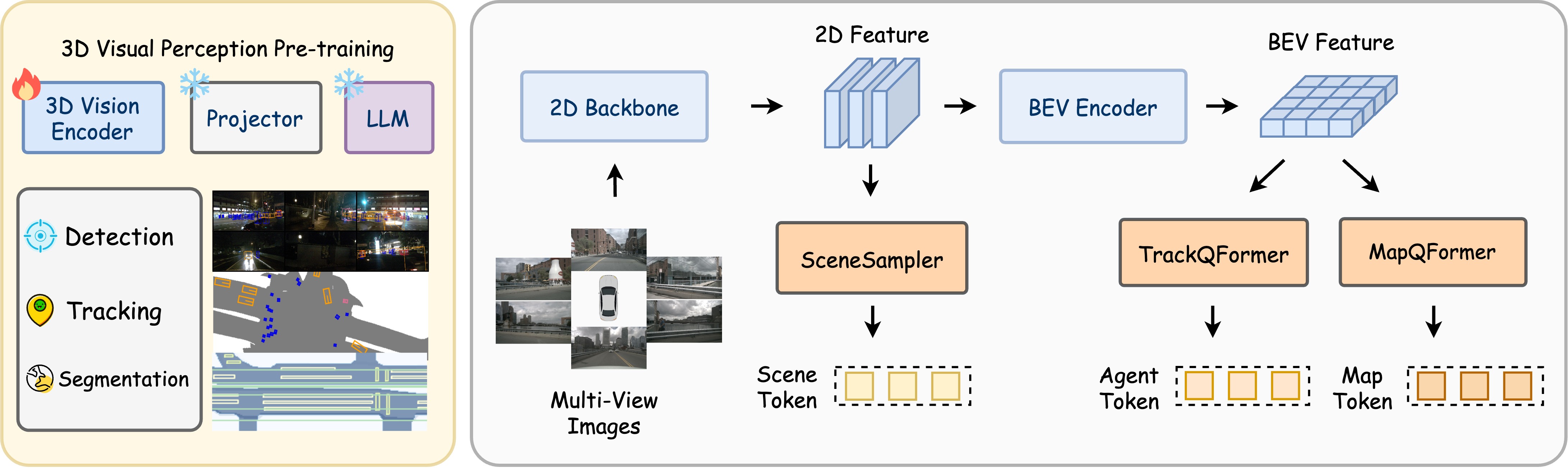}
    \caption{Overview of the visual encoder pretraining stage. The 3D vision-centric pretraining tasks include 3D object detection, tracking, and BEV panoptic segmentation. The extracted features are projected into scene, agent, and map tokens for downstream vision-language alignment and planning.}
\label{fig:stage0_vis1}
\end{figure}

We adopt the perception training stage introduced in UniAD~\cite{hu2023_uniad,contributors2023_uniadrepo} to produce a robust visual encoder for the traffic environment. It is trained via vision-centric 3D perception tasks, including 3D object detection, tracking, and BEV panoptic segmentation. Compared to grounding via general vision-language grounding tasks, vision-centric tasks offer more structured and semantically instance-level supervision, which is critical for autonomous driving.

The visual encoder follows a multi-view, query-based architecture, where multi-view images are processed by a shared ResNet-101 backbone with FPN to extract multi-scale 2D features, which are aggregated into BEV representation using BEVFormer~\cite{bevformer}. A detailed summary of the visual module setting in OpenDriveVLA is provided in Table~\ref{tab:vision_encoder_summary}.

\begin{table}[h]
    \centering
    \resizebox{\linewidth}{!}{
    \begin{tabular}{lllll}
        \toprule
        \textbf{Module} & \textbf{Input} & \textbf{Output} & \textbf{Architecture} & \textbf{Configuration} \\
        \midrule
        2D Backbone & Multi-view images & 2D features & ResNet-101 + FPN & Output strides: \{1/8, 1/16, 1/32\} \\
        BEV Encoder & 2D features & BEV feature map & BEVFormer (6-layer encoder) & Hidden dim: 256; BEV: $200 \times 200$ \\
        Scene Sampler & 2D features & Global scene token & Adaptive max pooling & Output size: $(6, 3, 5)$ \\
        Agent Query Transformer & BEV features & Agent tokens & TrackQFormer (6-layer decoder) & 900 queries \\
        Map Query Transformer & BEV features & Map tokens & MapQFormer (6-layer decoder) & 300 queries (3 thing + 1 stuff) \\
        \bottomrule
    \end{tabular}}
    \caption{Summary of components in the 3D visual perception module.}
    \label{tab:vision_encoder_summary}
\end{table}

The vision-centric task is supervised using a combination of detection, tracking and segmentation losses following original setting:

\begin{equation}
\mathcal{L}_{\text{vis}} = \mathcal{L}_{\text{track}} + \mathcal{L}_{\text{map}} \tag{13}
\end{equation}

The tracking loss $\mathcal{L}_{\text{track}}$ combines focal classification loss and L1 bounding box regression, optimized via a Hungarian matching algorithm. The map segmentation loss $\mathcal{L}_{\text{map}}$ includes classification, bounding box, mask, and IoU losses over both thing and stuff classes. Map supervision follows Panoptic SegFormer~\cite{li2022panopticsegformerdelvingdeeper}. This vision-centric supervision provides dense and structured grounding signals, enabling the encoder to learn spatially precise and semantically aligned features. 

% The resulting visual representations serve as input for robust structured token extraction in OpenDriveVLA.

\subsubsection{Structural Token Extraction}

Based on multi-scale 2D features $\mathbf{F}_{2D}$ from multi-view images and  aggregated BEV representation $\mathbf{F}_{bev}$, OpenDriveVLA retains the structured perception architecture of UniAD and extracts semantically meaningful tokens through three query-based modules. These modules encode complementary aspects of the driving scene and serve as a compact interface for downstream alignment and planning.

\textbf{Agent QueryTransformer:}
TrackQFormer decodes dynamic object-level semantics from the BEV feature map $f_{bev} \in \mathbb{R}^{200 \times 200 \times D}$ using learnable queries. Each token encodes an individual agent’s spatial location, category, and motion trajectory. We retain the top-$N_a$ tokens filtered by confidence threshold: $\{v_{\mathrm{agent}}^i\}{i=1}^{N_a} = \mathcal{Q}_{\mathrm{agent}}(f_{bev})$.
To improve efficiency and robustness, we filter out low-confidence agent predictions based on detection scores. This selective mechanism reduces the number of input visual tokens, leading to faster downstream processing and a more focused representation. Moreover, it helps mitigate hallucination by suppressing uncertain or noisy detections from the vision input source.

\textbf{Map QueryTransformer:}
MapQFormer focuses on extracting static structural elements from the BEV features via separate decoder heads for thing and stuff categories. It produces up to $N_m$ map tokens, each encoding elements such as lane dividers, road boundaries, and drivable areas:
$\{v_{\mathrm{map}}^j\}{j=1}^{N_m} = \mathcal{Q}_{\mathrm{map}}(f_{bev})
$.

\textbf{Global Scene Sampler:}
To capture holistic scene-level context, we apply adaptive max pooling over the 2D features $\mathbf{F}_{2D} \in \mathbb{R}^{6 \times 256 \times H \times W}$ from six camera views, compressing them into $(3 \times 5)$ spatial grids per view.  These tokens encode global contextual information, including weather, lighting, scene layout, and traffic flow, which are difficult to infer solely from BEV-based modules. By abstracting high-level semantics from raw visual features, the scene tokens serve as a complementary source of information to the agent and map tokens, providing redundancy and enhancing robustness for downstream alignment and planning:
$v_{\mathrm{scene}} = \mathcal{Q}_{\mathrm{scene}}(f_{2D}) \in \mathbb{R}^{90 \times D}
$.

Together, three token extractors provide a compact and structural encoding of the traffic environment, covering dynamic agents, static infrastructure, and global context:
$V_{\text{env}} = \{V_{\text{scene}}, V_{\text{agent}}, V_{\text{map}}\}$.

\subsubsection{Large Language Action Module}

\begin{table}[h]
\centering
\scriptsize
\resizebox{0.68\textwidth}{!}{%
\begin{tabular}{ll}
\toprule
\textbf{Token / Index} & \textbf{Description} \\
\midrule
\multicolumn{2}{l}{\textit{Placeholder Token Indices}} \\
\texttt{IMAGE\_TOKEN\_INDEX } & Placeholder for global image token \\
\texttt{SCENE\_TOKEN\_INDEX} & Placeholder for scene-level visual token \\
\texttt{TRACK\_TOKEN\_INDEX} & Placeholder for agent-level track token \\
\texttt{MAP\_TOKEN\_INDEX} & Placeholder for map-level token \\
\texttt{OBJECT\_TOKEN\_INDEX} & Reserved placeholder for object token \\
\midrule
\multicolumn{2}{l}{\textit{High-Level Markers}} \\
\texttt{<SCENE>}, \texttt{<TRACK>}, \texttt{<MAP>} & Denote visual token segments \\
\texttt{<EGO>} & Textualized ego-vehicle state \\
\texttt{<COMMAND>} & Driving command or query string \\
\texttt{<trajectory>} & Start of autoregressive trajectory output \\
\midrule
\multicolumn{2}{l}{\textit{Token Delimiters (Start / End Wrappers)}} \\
\texttt{<scene\_start>}, \texttt{<scene\_end>} & Scene token span delimiters \\
\texttt{<track\_start>}, \texttt{<track\_end>} & Track token span delimiters \\
\texttt{<map\_start>}, \texttt{<map\_end>} & Map token span delimiters \\
\texttt{<ego\_start>}, \texttt{<ego\_end>} & Ego state string delimiters \\
\texttt{<command\_start>}, \texttt{<command\_end>} & Driving command delimiters \\
\texttt{<traj\_start>}, \texttt{<traj\_end>} & Generated trajectory delimiters \\
\midrule
\multicolumn{2}{l}{\textit{Optional QA Format Tokens}} \\
\texttt{<question\_start>}, \texttt{<question\_end>} & Input question delimiters \\
\texttt{<answer\_start>}, \texttt{<answer\_end>} & Output answer delimiters \\
\bottomrule
\end{tabular}}
\caption{Extended special tokens and placeholder indices used in OpenDriveVLA's LLM tokenizer.}
\label{tab:special_tokens}
\end{table}

We adopt Qwen2.5-Instruct~\cite{qwen2.5} as the pre-trained LLM backbone for generating structured driving actions. To balance model capacity and computational efficiency, we evaluate three variants with 0.5B, 3B, and 7B parameters. Our implementation builds upon the LLaVA NeXT framework~\cite{liu2024llavanext}, which enables the language model to perform structured cross-modality reasoning and autoregressive action generation.

To support structured multimodal OpenDriveVLA input, we extend the Qwen2.5 tokenizer with a set of special tokens and token indices, as summarized in Table~\ref{tab:special_tokens}. These tokens formulate inputs into discrete semantic segments, allowing the LLM to differentiate between scene-level context, dynamic agents, static maps, ego vehicle state, and high-level driving instructions. The input sequence passed to the LLM is structured as:
\begin{center}
\texttt{<SYSTEM><SCENE><TRACK><MAP><EGO><COMMAND>}
\end{center}

During input construction, \texttt{<SCENE>}, \texttt{<TRACK>}, and \texttt{<MAP>} are replaced with projected visual tokens, while \texttt{<EGO>} and \texttt{<COMMAND>} are filled with formatted textual strings. 

\clearpage

\subsection{Prompting Techniques}

\subsubsection{System Prompt}

The system prompt follows the BEV coordinate conventions introduced in GPT Driver~\cite{mao2024gptdriver}, while specifying the model role, driving objectives, and output format. It guides the LLM to perform perception, reasoning, and trajectory generation in a unified manner, with optional user interaction. The system prompt is prepended to all inputs during both training and inference to ensure consistent task framing and instruction following.

\begin{tcolorbox}[title=System Prompt, colback=gray!5!white, colframe=gray!75!black, label={lst:system_prompt}]
    \begin{lstlisting}[basicstyle=\ttfamily\footnotesize,breaklines=true]
    <|im_start|>system
    You are Open-DriveVLA, an advanced vision-language driving model. Your core responsibilities include safe trajectory planning and interpretable decision-making. You generate collision-free driving plans while providing clear, logical explanations for user queries.
    
    Context:
     - Coordinates: X-axis is pointing to the right, and Y-axis is pointing to the front. You're at point (0,0). All coordinates are in meters.
     - Objective: Generate a 3-second safe driving plan consisting of 6 waypoints, one every 0.5 seconds. Provide logical responses to user queries.
    
    Task:
    - Perception & Prediction: Analyze the driving environment using visual data. Identify road users and hazards and predict their motion.
    - Thought Process: Determine critical objects and assess potential hazards. Consider road constraints and traffic rules.
    - Trajectory Planning: Define the driving objective. Generate a safe, feasible 3-second route consisting of 6 waypoints.
    - Explainability & User Interaction: If the user asks a question, provide a clear and logical response.
    
    Output Format:
    1. Trajectory (MOST IMPORTANT):
      - Format: <traj_start>[(x1,y1),(x2,y2),(x3,y3),(x4,y4),(x5,y5),(x6,y6)]<traj_end>
    2. User Question Response (OPTIONAL):
      - Format: <answer_start> Answer to the user's question <answer_end>
    \end{lstlisting}
\end{tcolorbox}

\subsubsection{Prompts for Hierarchical Feature Alignment}
Each type of visual token is associated with a separate captioning prompt. These prompts instruct the LLM to generate textual descriptions based on the corresponding visual input segment.

\begin{tcolorbox}[title=Instance Caption, colback=gray!5!white, colframe=gray!75!black, label={lst:Prompt_for_Instance_caption}]
    \begin{lstlisting}[basicstyle=\ttfamily\footnotesize,breaklines=true]
    Please provide a caption and the BEV coordinate for the following object: <track_start><OBJECT><track_end>

    \end{lstlisting}
\end{tcolorbox}

\begin{tcolorbox}[title=Map Caption, colback=gray!5!white, colframe=gray!75!black, label={lst:Prompt_for_map_caption}]
    \begin{lstlisting}[basicstyle=\ttfamily\footnotesize,breaklines=true]
    Please provide a caption for the following map: <map_start><MAP><map_end>
    \end{lstlisting}
\end{tcolorbox}

\begin{tcolorbox}[title=Scene Caption, colback=gray!5!white, colframe=gray!75!black, label={lst:Prompt_for_scene_caption}]
    \begin{lstlisting}[basicstyle=\ttfamily\footnotesize,breaklines=true]
    Please provide a caption for the following scene: <scene_start><SCENE><scene_end>
    \end{lstlisting}
\end{tcolorbox}

\subsubsection{Prompts for Driving Question Answering}
All VQA datasets adopt a consistent structured prompt format, integrating scene, agent, map, ego state, and historical trajectory components. Question Placeholders are replaced with dataset-specific content during training and evaluation.

\begin{tcolorbox}[title=Driving Question Answering Prompt, colback=gray!5!white, colframe=gray!75!black, label={lst:Prompt_for_QA}]
    \begin{lstlisting}[basicstyle=\ttfamily\footnotesize,breaklines=true]
      Scene information: <scene_start><SCENE><scene_end>\nObject-wise tracking information: <track_start><TRACK><track_end>\nMap information: <map_start><MAP><map_end>\nEgo states: - Velocity (vx,vy): <Velocity Placeholder> - Heading Angular Velocity (v_yaw): <Angular Velocity Placeholder> - Acceleration (ax,ay): <Acceleration Placeholder> - Can Bus: <Can Bus Placeholder> - Heading Speed: <Speed Placeholder> - Steering: <Steering Placeholder>\nHistorical trajectory (last 2 seconds): <Trajectory Placeholder> \nPlease answer the following question: <Question Placeholder>

    \end{lstlisting}
\end{tcolorbox}

\subsubsection{Prompts for Agent-Env-Ego Interaction}
This prompt is used in Stage 2.5-modeling agent-environment-ego interactions. The input includes structured visual context, historical ego trajectory, and a queried target object, and the model is instructed to predict the future motion of the agent.

\begin{tcolorbox}[title=Driving Question Answering Prompt, colback=gray!5!white, colframe=gray!75!black, label={lst:Prompt_for_agent_env_ego}]
    \begin{lstlisting}[basicstyle=\ttfamily\footnotesize,breaklines=true]
      <scene_start><SCENE><scene_end>\nObject-wise tracking information: <track_start><TRACK><track_end>\nMap information: <map_start><MAP><map_end>\nEgo Vehicle Token: <trajectory>\nPlease predict relative motion trajectory for the following object: <track_start><OBJECT><track_end>


    \end{lstlisting}
\end{tcolorbox}

\subsubsection{Prompts for Trajectory Planning Tuning}
This prompt is used in the training stage 3 for trajectory planning tuning, where the model is supervised to generate a 3-second driving plan based on structured multi-modal context.

\begin{tcolorbox}[title=Driving Question Answering Prompt, colback=gray!5!white, colframe=gray!75!black, label={lst:Prompt_for_driving_question}]
    \begin{lstlisting}[basicstyle=\ttfamily\footnotesize,breaklines=true]
      Scene information: <scene_start><SCENE><scene_end>\nObject-wise tracking information: <track_start><TRACK><track_end>\nMap information: <map_start><MAP><map_end>\nEgo states: - Velocity (vx,vy): <Velocity Placeholder> - Heading Angular Velocity (v_yaw): <Angular Velocity Placeholder> - Acceleration (ax,ay): <Acceleration Placeholder> - Can Bus: <Can Bus Placeholder> - Heading Speed: <Speed Placeholder> - Steering: <Steering Placeholder>\nHistorical trajectory (last 2 seconds): <Trajectory Placeholder>\nMission goal: <Command Placeholder>\nPlanning trajectory: <trajectory>

    \end{lstlisting}
\end{tcolorbox}

\subsection{Training and Inference Details}

\subsubsection{Training Configuration.} OpenDriveVLA is trained on 4 NVIDIA H100 GPUs with a per-GPU batch size of 1. The full training process takes approximately two days for the 0.5B variant. All training stages use mixed-precision (bf16) and gradient checkpointing to improve memory efficiency and speed. The 2D vision backbone is frozen during the final end-to-end stage. LLM parameters are fully tuned unless specified otherwise.

\begin{table}[h]
\centering
\resizebox{0.90\textwidth}{!}{
\begin{tabular}{c|c|cccc}
\toprule
 & & \textbf{Stage 1} & \textbf{Stage 2} & \textbf{Stage 2.5} & \textbf{Stage 3}\\
\midrule
\multirow{6}{*}{\textbf{Train}} 
& Tunable parts & projector & projector,LLM & projector,LLM & Full model (except 2D encoder)\\
& Trainable Params (MB) & 3.1 & 496.9 & 496.9 & 552.6 \\
& Per-GPU batch size & 1 & 1 & 1 & 1\\
& GPUs & 4 & 4 & 4 & 4 \\
& LR ($\psi_{\text{vision}}$) & – & - & - & $1\times10^{-5}$\\
& LR ($\{\theta_{\text{proj/LLM}}\}$) & $1\times10^{-4}$ & $1\times10^{-5}$ & $1\times10^{-5}$ & $1\times10^{-5}$\\
% & Weight decay & 0 & 0 & 0 & 0\\
% & Warm-up ratio & 0.03 & 0.03 & 0.03 & 0.03\\
% & Scheduler & cosine & cosine & cosine & cosine\\
% & Frames upper-bound & – & 32 & 32 & 32\\
& Epochs & 1 & 1 & 1 & 1\\
\bottomrule
\end{tabular}}
\caption{Multi-stage training hyperparameters of OpenDriveVLA-0.5B.}
\label{tab:training_pipeline}
\end{table}

% \begin{table}[ht]
% \centering
% \caption{Training Overview of the multi-stage training details of OpenDriveVLA, including data, model, and parameters.}
% \resizebox{0.95\textwidth}{!}{ 
% \begin{tabular}{c|c|ccccc}
%     \toprule
%     & & \textbf{Stage 0} & \textbf{Stage 1} & \textbf{Stage 2} & \textbf{Stage 2.5} & \textbf{Stage 3} \\ 
%     \midrule

%     \multirow{4}{*}{\textbf{Training}} 
%     & Trainable
%     & Vision Tower
%     & Proj 
%     & Proj, LLM 
%     & Proj, LLM 
%     & Full Model \\ 
%     & 0.5B Size (MB) 
%     & 
%     & 3.1 
%     & 496.9 
%     & 496.9  
%     & 552.6 \\ 
%     & 3B Size (MB) 
%     & 
%     & 14.2 
%     & 3099.6 
%     & 3099.6  
%     & 3155.3 \\ 
%     & 7B Size (MB) 
%     &
%     & 41.3
%     & 7654.2 
%     & 7654.2 
%     & 7709.9 \\ 
%     \midrule
%     \multirow{2}{*}{\textbf{Inference}}
%     &Speed
%     &
%     & \\
%     &
%     &
%     &
    
%     \\

%     \bottomrule
% \end{tabular}
% }
% \label{tab:training_pipeline}
% \end{table}

\subsubsection{Inference Efficiency} Table~\ref{tab:inference_perf} reports the inference performance of OpenDriveVLA across three LLM scales (0.5B, 3B, 7B) under bf16 precision on a single NVIDIA A100 GPU. Evaluation is conducted on the NuScenes trajectory validation set with 6019 samples.

\begin{table}[h]
\centering
\resizebox{0.83\linewidth}{!}{
\begin{tabular}{l|c|c|c|c|c}
\toprule
Model & LLM & GPUs & Speed (Sample/s) & Latency (s) & Max VRAM (GB) \\ \midrule
0.5B & Qwen2.5-0.5B-Instruct & 1 & 0.74 & 1.36 & 1.56 \\
3B   & Qwen2.5-3B-Instruct   & 1 & 0.54 & 1.85 & 7.35 \\
7B   & Qwen2.5-7B-Instruct   & 1 & 0.57 & 1.74 & 17.15 \\
\bottomrule
\end{tabular}
}
\caption{Inference details of OpenDriveVLA under BF16 precision on a single NVIDIA A100 GPU.}
\label{tab:inference_perf}
\end{table}

\clearpage
% \section{Datasets}
% \clearpage
\section{Datasets Details}

\subsection{Dataset Overview}
We utilize a curated set of multimodal driving datasets derived from nuScenes \cite{nuScenes} to support the multi-stage training of OpenDriveVLA, covering object-level captioning, visual question answering, scene description, and decision reasoning. An overview of the datasets used in Stage 1 and Stage 2 of OpenDriveVLA training is provided in Table~\ref{tab:qa_dataset_details}.

\begin{table}[h]
    \centering
    \resizebox{0.97\linewidth}{!}{
    \begin{tabular}{lcccc}
        \toprule
        \textbf{Dataset} & \textbf{\#Train} & \textbf{\#Val} & \textbf{Annotation Types} & \textbf{Type} \\
        \midrule
        TOD3Cap      & 1.89M & 410K & Object Caption: appearance, motion, relationships & Dense Captioning \\
        nuScenes-QA   & 376K  & 83K  &  existence, counting, object, status, comparison & Visual Question Answering \\
        nuCaption     & 348K  & 72K  & Scene Caption: layout, agents, hazards & Scene Description \\
        nuX         & 28K   & 6K   & Driving Decision Justification & Reasoning \& Narration \\
        \bottomrule
    \end{tabular}}
    \caption{VQA Datasets in OpenDriveVLA Stage 1/2 training, with sample counts, annotation, and types.}
    \label{tab:qa_dataset_details}
\end{table}

TOD3Cap \cite{TOD3Cap}: TOD3Cap introduces the task of object-centric dense captioning in 3D driving scenes. It provides 2.3M human-verified natural language descriptions for over 64K objects across 850 nuScenes \cite{nuScenes} scenes, covering appearance, motion, context, and inter-object relations. Each caption captures fine-grained semantics including what, where, how, and why, facilitating object-level alignment between 3D perception and language. 

nuScenes-QA \cite{nuscenesqa}: nuScenes-QA is a large-scale visual question answering benchmark tailored for autonomous driving. It contains 460K question–answer pairs over 34K multimodal driving scenes with synchronized images and LiDAR. Questions are generated using scene graphs and structured templates, spanning reasoning types such as existence, counting, attribute queries, spatial relations, and comparisons.

nuCaption \cite{nucaption}: nuCaption is a 3D scene captioning dataset constructed from nuScenes. It comprises both image-text and LiDAR-text pairs, with both global and viewpoint-specific captions describing traffic layout, object interactions, and potential risks. By aligning 3D spatial representations with language, nuCaption allows 3D captioning and scene-level reasoning.

nuX \cite{hintad-nux}: nuX is a human-annotated explanation dataset designed for interpretable autonomous driving. For each keyframe, it provides natural language explanations combining factual narration (what is happening) and causal reasoning (why it is happening), grounded in the outputs of perception, prediction, and planning modules. It supports the development of driving models with aligned and transparent decision-making with textual interpretability.

\subsection{Dataset Sample Visualization}

\begin{table}[h]
\centering
\resizebox{\textwidth}{!}{ 
\begin{tabular}{lccccc}
    \toprule
    & & \textbf{Stage 1} & \textbf{Stage 2} & \textbf{Stage 2.5} & \textbf{Stage 3} \\ 
    \midrule
    % \multirow{1}{*}{} 
    & Annos 
    % & Bbox \&Seg \&Track
    & 2D/3D Caption, Scene Description
    & Driving QA 
    & Trajectory 
    & Trajectory \\ 
    & Task 
    % & Vision-centric Training 
    &  Hierarchical Feature Alignment & Driving Instruction Tuning & Agent-Env-Ego Interaction &  Trajectory Planning\\
    & Source
    % & nuScenes 
    & TOD3Cap, nuScenes, nuCaption
    & nuCaption, nuS-QA, nuX 
    & nuScenes
    & nuScenes
    \\
    
    & \#Samples 
    & 536k 
    & 566k
    & 459k 
    & 28k \\ 
    \bottomrule

\end{tabular}
}
\caption{Overview of the Dataset for multi-stage training pipeline of OpenDriveVLA, detailing the specific tasks, annotation types, data sources, and number of training samples.}
\label{tab:dataset_overview}
\end{table}

% \begin{table}[h]
%     \centering
%     \scriptsize
%     \begin{tabular}{lccc}
%         \toprule
%         Dataset & Training Samples & Validation Samples & Annotation Types \\
%         \midrule
%         TOD3Cap   & 1,890k & 410k & Appearance, Motion, Environment, Relationship \\
%         nuScenes-QA & 376k & 83k & Existence, Counting, Query-Object, Query-Status, Comparison \\
%         nuCaption & 348k & 72k & Scene Description \\
%         nuX & 28k & 6k & Perception (Narration), Reasoning \\
%         \bottomrule
%     \end{tabular}
%     \caption{Statistics of question-answer datasets used for stage 2 and 3 of the OpenDriveVLA training pipeline.}
%     \label{tab:qa_dataset_details}
% \end{table}

% \subsubsection{Visualization of Hierarchical Feature Alignment}
An overview of the multi-stage training data is provided in Table~\ref{tab:dataset_overview}, covering task types, annotations, data sources, and sample counts. We provide visualization examples of gannotations we used during the hierarchical feature alignment training process of OpenDriveVLA, as shown in Figure~\ref{fig:stage1_vis1} and Figure~\ref{fig:stage1_vis2}.

% Instance-level captions are shown in green and describe individual traffic participants such as vehicles and pedestrians, including their appearance, motion, spatial relation to the ego car, and corresponding BEV coordinates. Map-level captions, shown in yellow, summarize HD map content for each sample, covering static infrastructure elements like lane boundaries, dividers, and crosswalks. Scene-level captions are shown in red and provide holistic natural language descriptions that integrate information across all camera views to capture the overall layout, dynamics, and situational context of the scene.

\begin{figure}[h]
    \centering
    \includegraphics[width=0.78\linewidth]{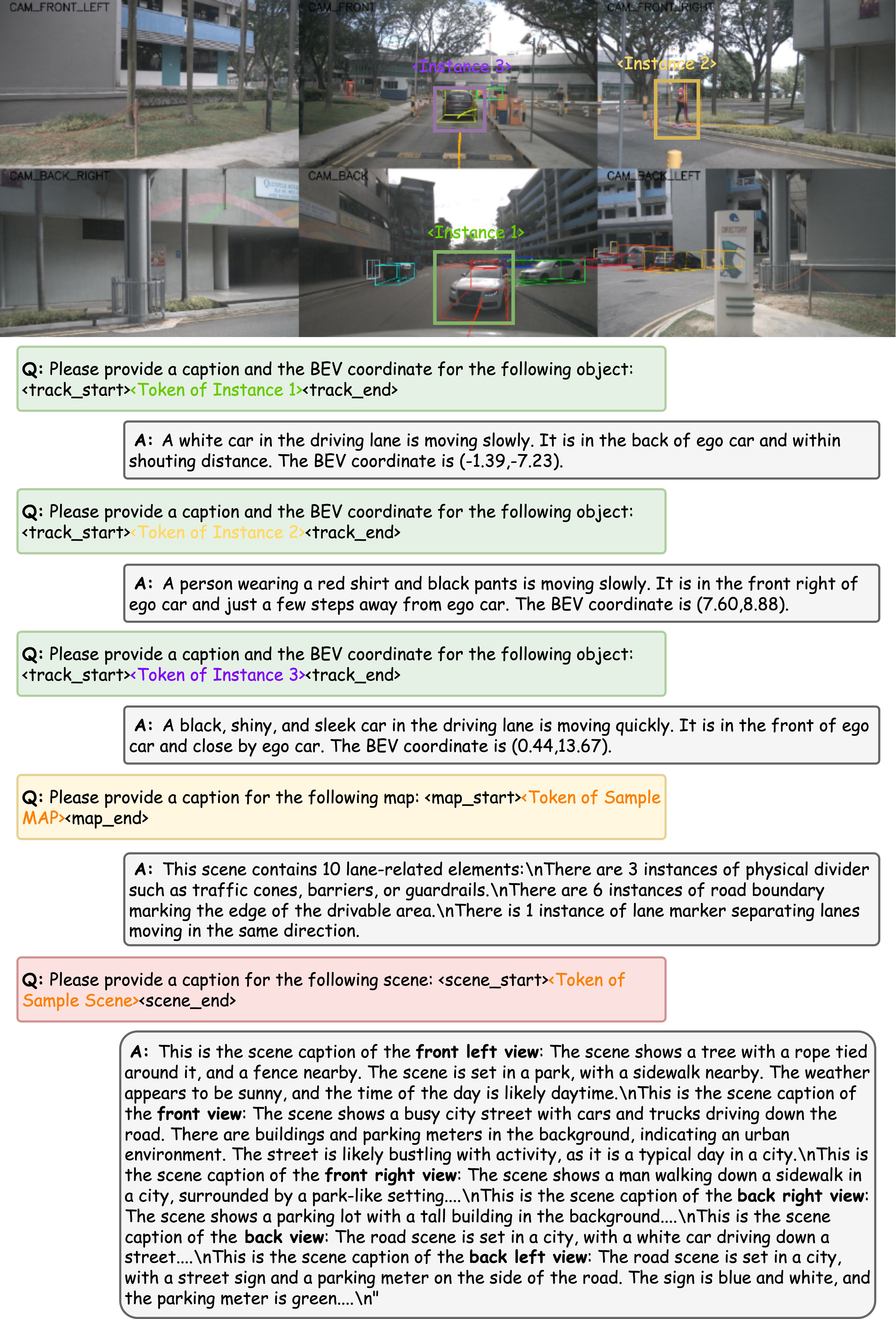}
    \caption{Example 1 of Stage 1 dataset visualization from nuScenes Sample 90. Green boxes show instance token captions; Yellow boxes indicate map token captions; Red boxes represent scene token captions.}
    \label{fig:stage1_vis1}
\end{figure}

\begin{figure}[h]
    \centering
    \includegraphics[width=0.78\linewidth]{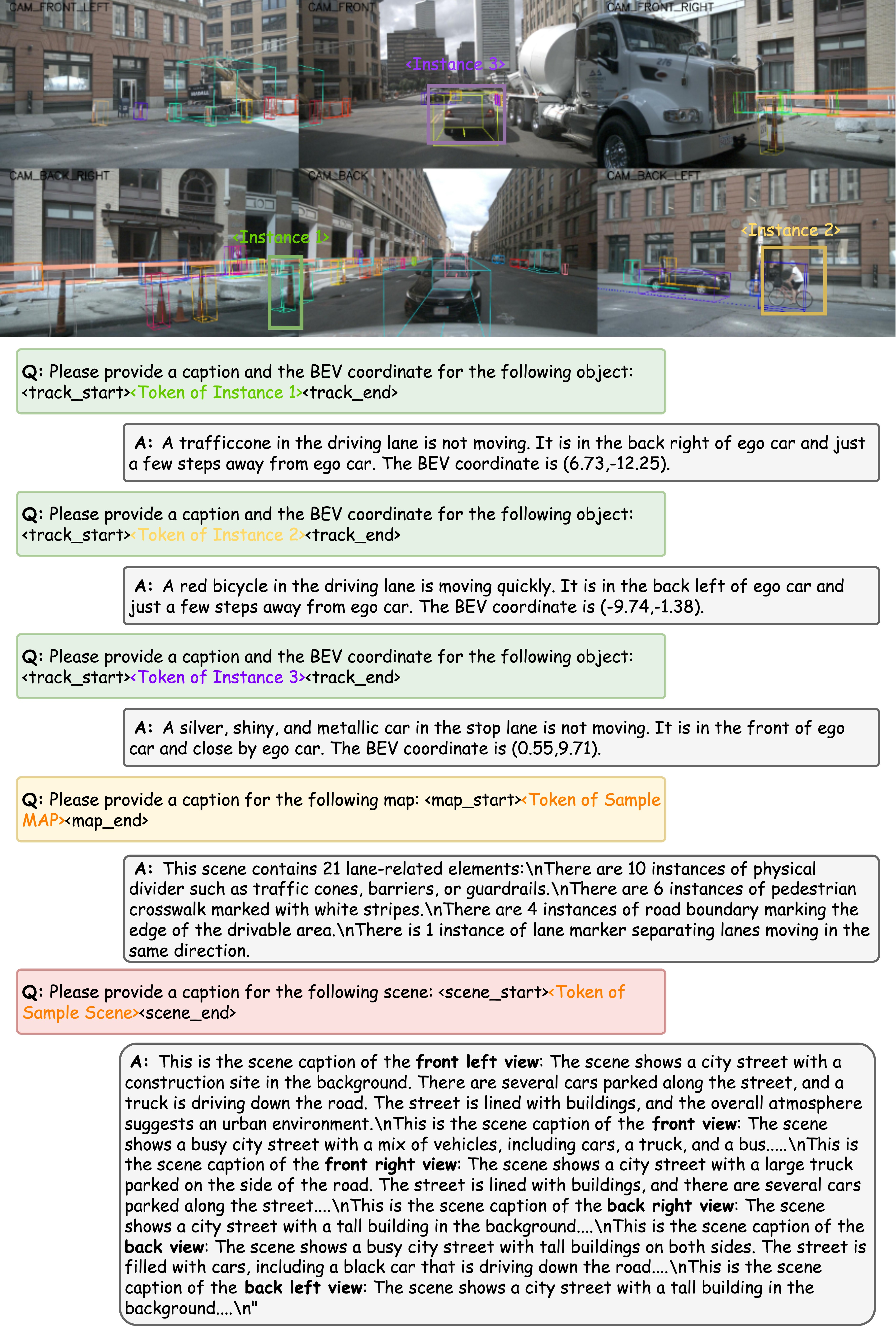}
    \caption{Example 2 of Stage 1 dataset visualization from nuScenes Sample 3287. Green boxes show instance token captions; Yellow boxes indicate map token captions; Red boxes represent scene token captions.}    
    \label{fig:stage1_vis2}
\end{figure}

\clearpage
\clearpage
% \section{Additional Results}

% ##################################################
\section{Results and Discussions}

\subsection{Results of Driving Question Answering}

Figure~\ref{fig:stage2_vis1} and Figure~\ref{fig:stage2_vis2} show representative examples of OpenDriveVLA's responses to diverse driving-related questions drawn from three datasets: nuScenes-QA, nuCaption, and nuX. These qualitative results highlight the model’s multi-level reasoning capabilities across perception, commonsense understanding, and contextual decision-making.

In Figure~\ref{fig:stage2_vis1}, the model provides a narration of the ego vehicle’s behavior while approaching and passing over a speed bump, accompanied by a causal explanation grounded in the surrounding scene. This demonstrates the model’s ability to generate coherent natural language outputs that reflect its interpretation of driving decisions. However, in the scene-level description, the model hallucinates that there are no pedestrians visible in the right-front view, despite the presence of one pedestrian in the image.

\begin{figure}[h]
    \centering
    \includegraphics[width=0.85\linewidth]{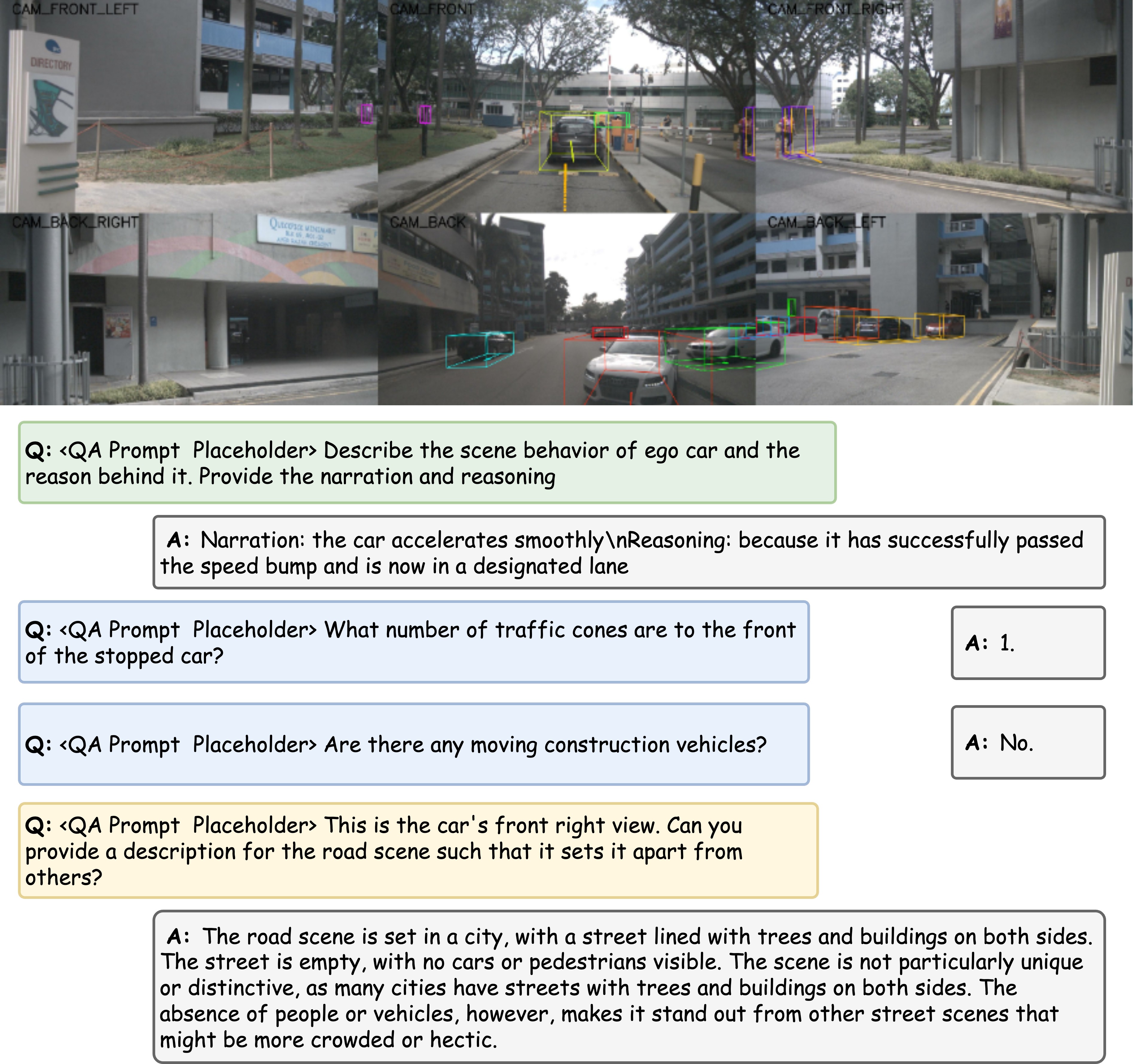}
    \caption{Example 1 of Stage 2 dataset visualization from nuScenes Sample 6801. Green boxes show nuX dataset predictions; blue boxes are from nuScenes-QA dataset, while yellow boxes are from nuCaption.}

    \label{fig:stage2_vis1}
\end{figure}

\clearpage

Figure~\ref{fig:stage2_vis2} depicts a more complex urban scenario involving multiple interacting agents near an intersection. The model identifies the ego vehicle’s maneuver and provides a causal explanation grounded in its spatial context, correctly localizing the vehicle as it turns at the junction. In the scene-level description, the model can ground key static and dynamic elements, including the presence of a traffic light, pedestrians near a crosswalk, and parked vehicles along the roadside. However, the caption exhibits positional inaccuracy when referencing the camera view, particularly in the identification of the front view. This may be attributed to the model being trained with scene-level captions in a camera-agnostic fashion, without explicit supervision for distinguishing between different camera perspectives.

\begin{figure}[h]
    \centering
    \includegraphics[width=0.85\linewidth]{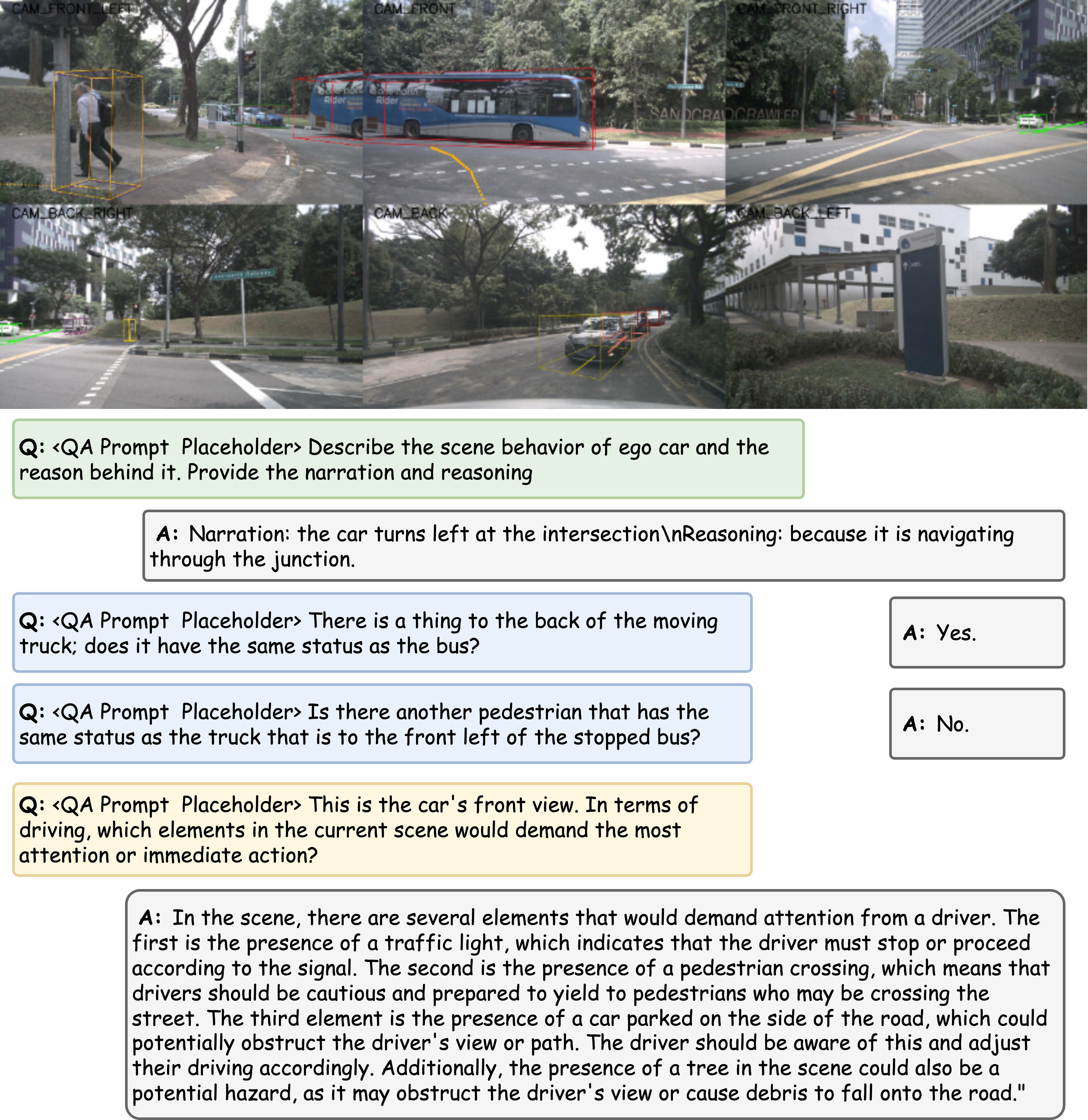}
    \caption{Example 2 of Stage 2 dataset visualization from nuScenes Sample 6801. Green boxes show nuX dataset predictions; blue boxes are from nuScenes-QA dataset, while yellow boxes are from nuCaption.}

    \label{fig:stage2_vis2}
\end{figure}

% \clearpage
\subsection{Results of Agent Motion Prediction}

Figure~\ref{fig:stage25_vis1} and Figure~\ref{fig:stage25_vis2} present qualitative results of OpenDriveVLA on the agent motion prediction task of Stage 2.5, where the model jointly reasons over agent trajectories, environment, and ego vehicle state. 
% These examples illustrate the model’s ability to generate spatially and temporally grounded motion predictions in complex urban scenes.

\begin{figure}[h]
    \centering
    \includegraphics[width=0.75\linewidth]{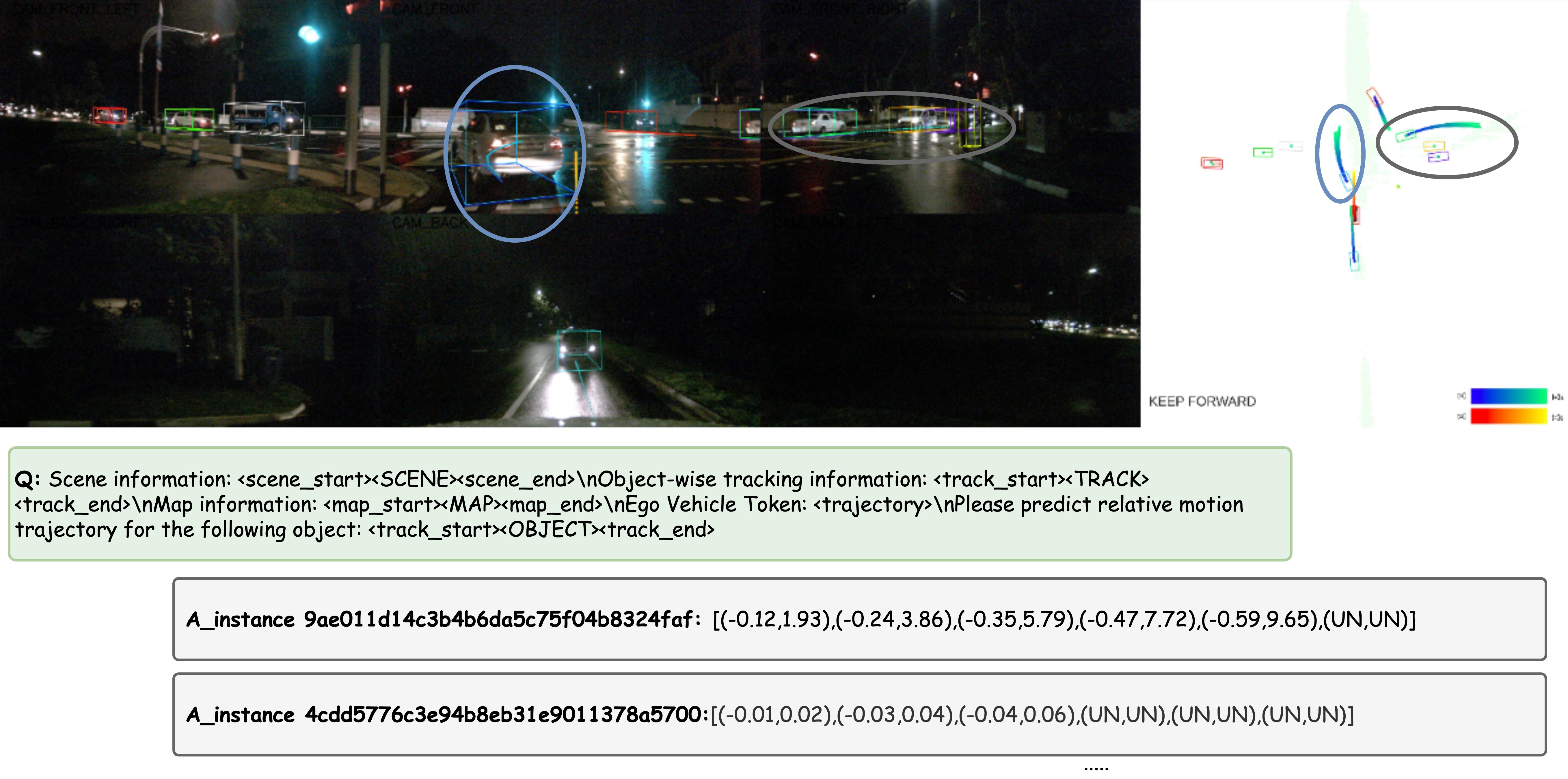}
    \caption{Example 1 of agent motion prediction result of stage 2.5 Agent-Environment-Ego interaction modeling on nuScenes validation set Sample 8587.}
    \label{fig:stage25_vis1}
\end{figure}

Figure~\ref{fig:stage25_vis1} depicts a night-time urban intersection with multiple static and dynamic agents under low-light conditions. The figure highlights two turning vehicles. The predicted motion for the white sedan on the right side of the scene is consistent with the lane orientation and road semantics. In contrast, the white vehicle ahead of the ego car, the model predicts a trajectory curving toward the right. Yet, given the surrounding road geometry and lane configuration, a left turn would be the more plausible maneuver.

\begin{figure}[h]
    \centering
    \includegraphics[width=0.75\linewidth]{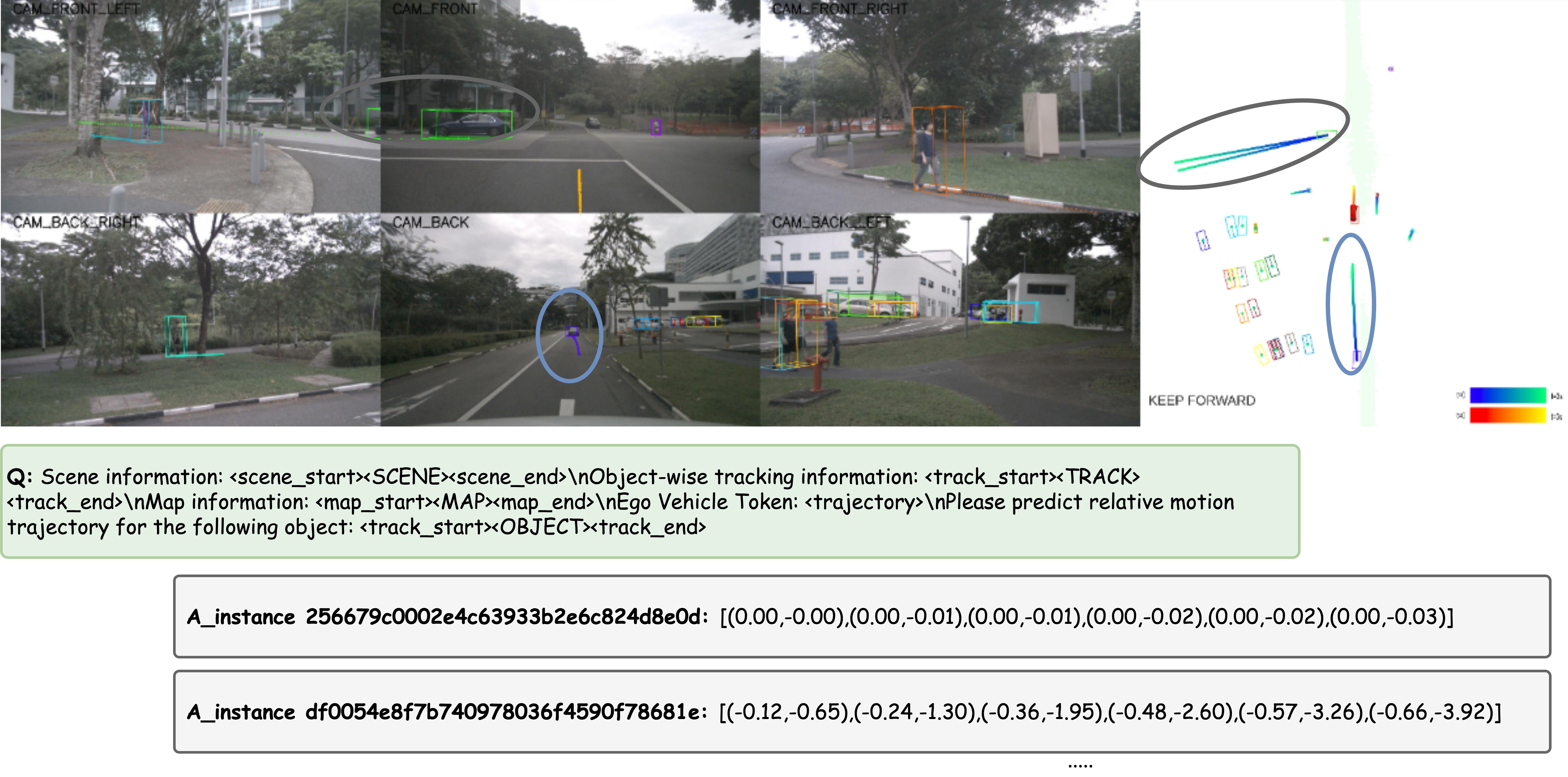}
    \caption{Example 2 of agent motion prediction result after stage 2.5 Agent-Environment-Ego interaction modeling on nuScenes validation set Sample 32496.}
    \label{fig:stage25_vis2}
\end{figure}

In Figure~\ref{fig:stage25_vis2}, the model predicts motion trajectories for multiple agents in a curved road scenario during daytime. Notably, the visualization highlights two predicted trajectories: one for a gray sedan in the front view, and another for a vehicle observed in the back view. Both predictions reflect distinct motion patterns consistent with the road layout and surrounding context. This example demonstrates the model’s capacity to capture plausible motion uncertainty and agent-specific intention under partially observable environments.

% \clearpage

% #############################

\subsection{Planning Results}

\subsubsection{Comparison with Prior Methods.}
Figure~\ref{fig:comparison} presents a qualitative comparison between the open-loop planning results of OpenDriveVLA and UniAD~\cite{hu2023_uniad} in a challenging narrow-road scenario with multiple parked vehicles. 
\begin{figure}[h!]
    \centering
    \begin{minipage}{0.98\linewidth}
        \centering
        \includegraphics[width=\textwidth]{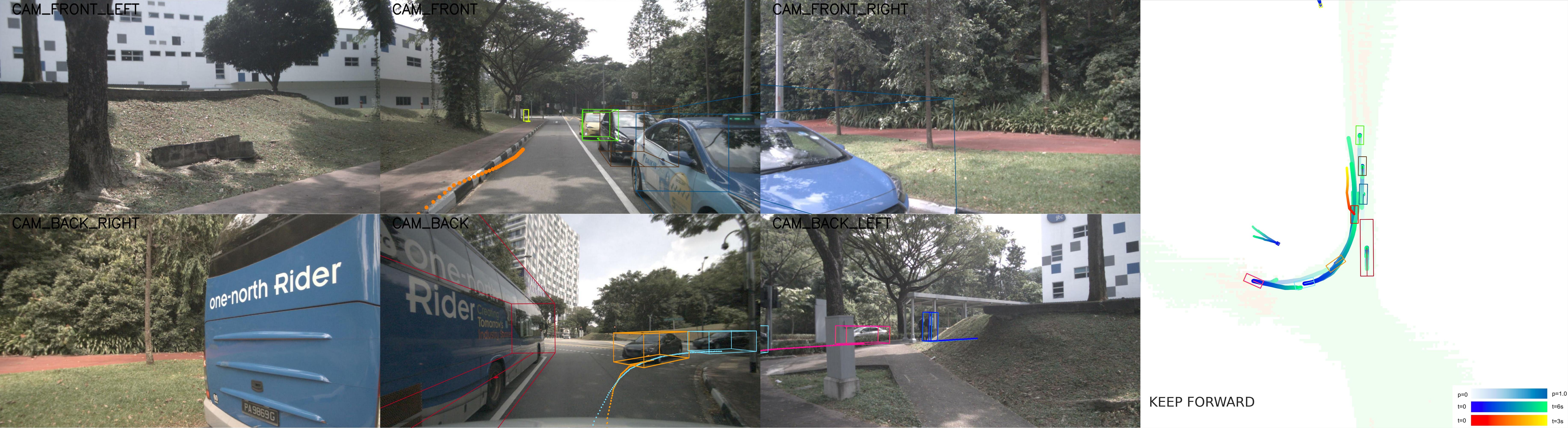}
        % \caption*{(a) UniAD \cite{hu2023_uniad}}
        \label{fig:comparison_uniad}
    \end{minipage}\par\medskip
    \begin{minipage}{0.98\linewidth}
        \centering
        \includegraphics[width=\textwidth]{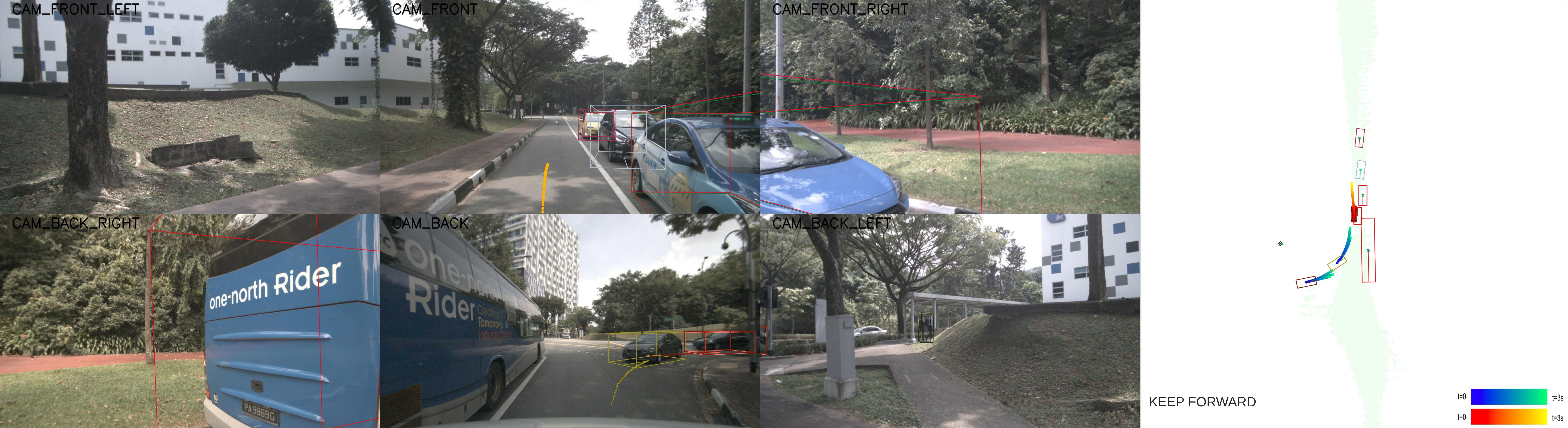}
        % \caption*{(b) OpenDriveVLA-7B (Ours)}
        \label{fig:comparison_DriveVLA}
    \end{minipage}

    \caption{Qualitative comparison of open-loop planning. Top: Planning results of UniAD \cite{hu2023_uniad}. Bottom: Planning results of OpenDriveVLA-7B (Ours). The agent motion prediction results are visualized after the agent-environment-ego interaction stage.}
    \label{fig:comparison}
\end{figure}

In this scenario, UniAD exhibits overly sensitive reactions to the parked vehicles on the right side of the road, resulting in unstable and zigzagging planned trajectories. In contrast, OpenDriveVLA generates smoother and more consistent motion plans that better follow the intended driving path while maintaining safe clearance from surrounding objects. This demonstrates a stronger ability to reason about scene semantics and generate robust and spatially grounded trajectories.

% \subsection{Limitations and Discussions}

\subsubsection{Qualitative Results of Driving Instruction Following.}
Figure~\ref{fig:stage3_compare1} and Figure~\ref{fig:stage3_compare2} illustrate the trajectory planning behavior of OpenDriveVLA under different conditional driving instructions, including left turn, right turn, and keep forward. 

Figure~\ref{fig:stage3_compare1} illustrates a complex urban intersection where the ground-truth instruction in nuScenes dataset is going forward. The predicted paths remain safe and well-aligned with the scene context under new driving instructions, demonstrating the model’s capacity to flexibly interpret high-level commands while maintaining collision-free behavior. Similarly, Figure~\ref{fig:stage3_compare2} shows another intersection scenario in a suburban environment. Despite differences in layout and visual context, OpenDriveVLA consistently adapts its planned motion to match the given instruction safely. These results demonstrate the instruction following ability of OpenDriveVLA across various environments and its robustness in instruction-conditioned planning.

\begin{figure}[h]
    \centering

    % Top image
    \includegraphics[width=0.98\linewidth]{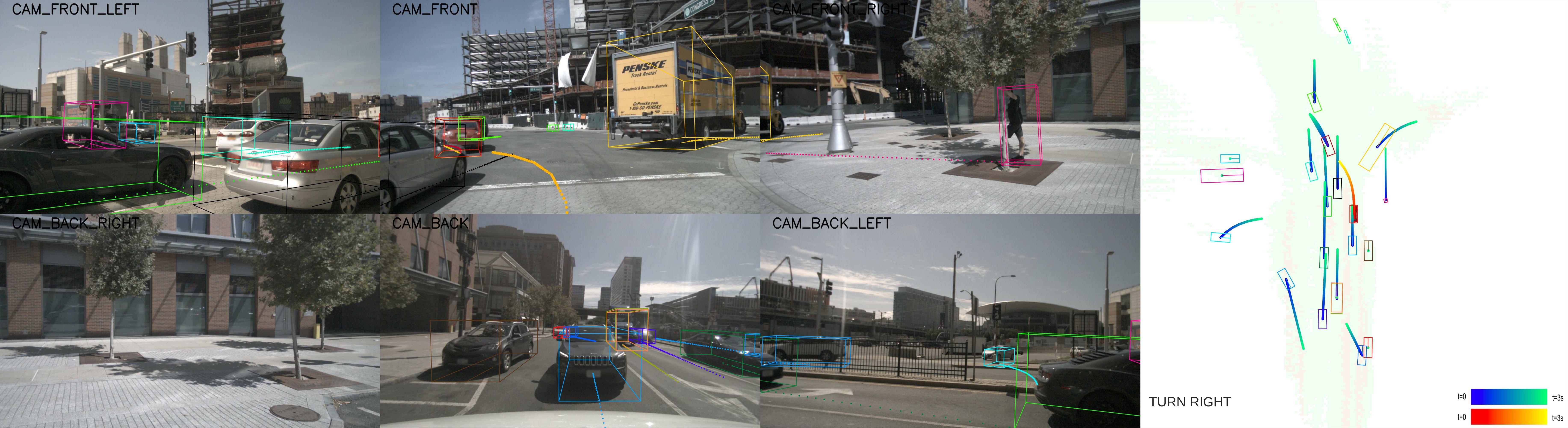}\\
    \small (a) OpenDriveVLA planning under instruction turning left

    \vspace{1.2em}

    % Middle image
    \includegraphics[width=0.98\linewidth]{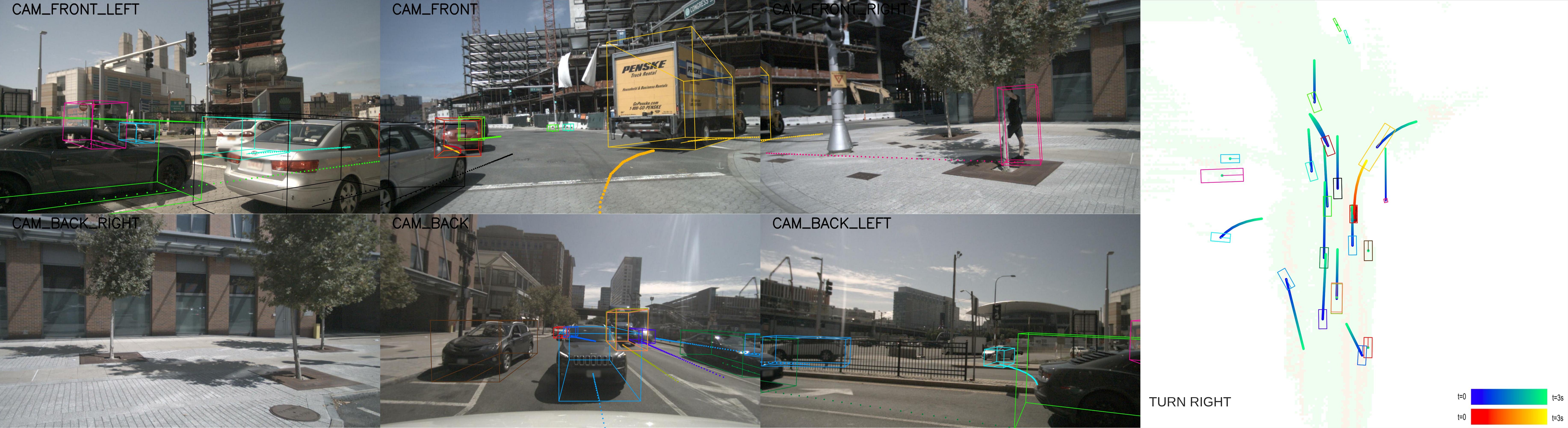}\\
    \small (b) OpenDriveVLA planning under instruction turning right

    \vspace{1.2em}

    % Bottom image
    \includegraphics[width=0.98\linewidth]{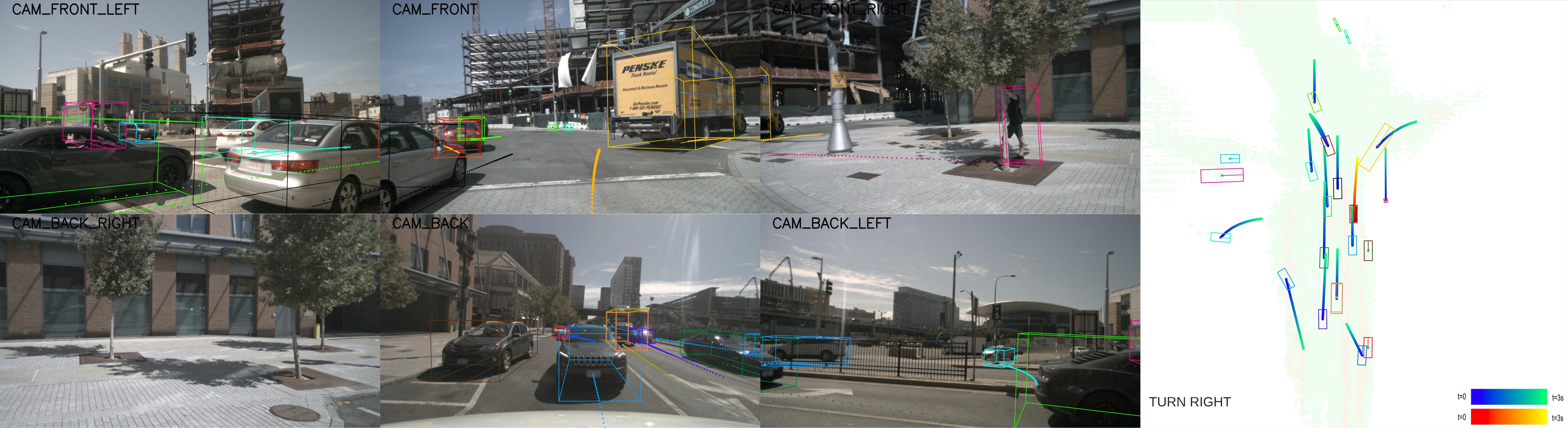}\\
    \small (c) OpenDriveVLA planning under ground-truth instruction: keep forward

    \caption{Example 1 of trajectory planning results of OpenDriveVLA after stage 3 with different driving instructions.}

    \label{fig:stage3_compare1}
\end{figure}

\begin{figure}[h]
    \centering

    % Top image
    \includegraphics[width=0.98\linewidth]{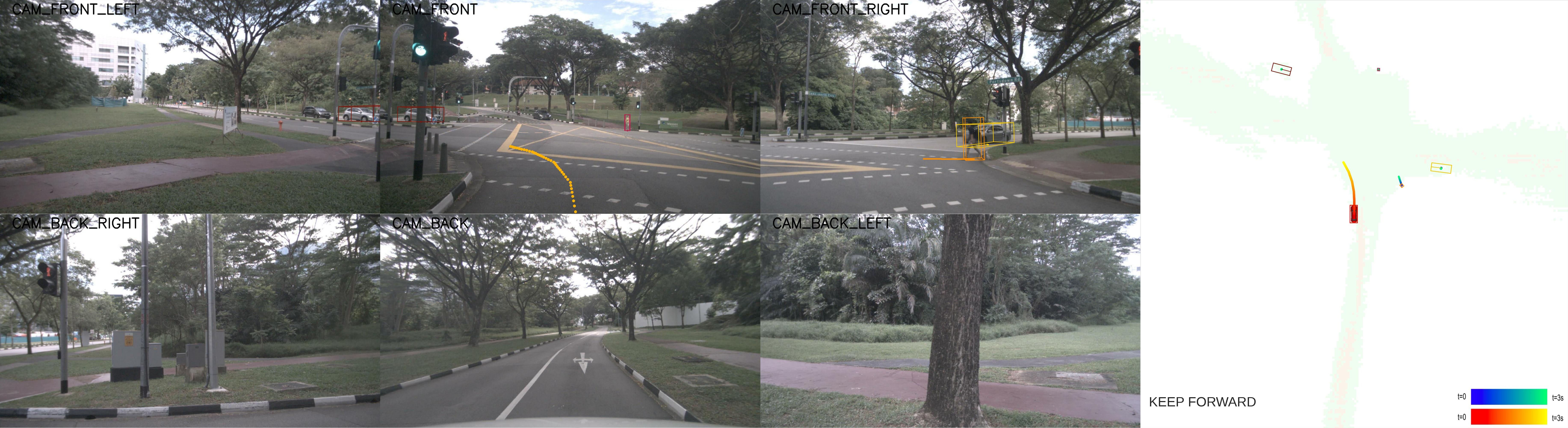}\\
    \small (a) OpenDriveVLA planning under instruction turning left

    \vspace{1.2em}

    % Middle image
    \includegraphics[width=0.98\linewidth]{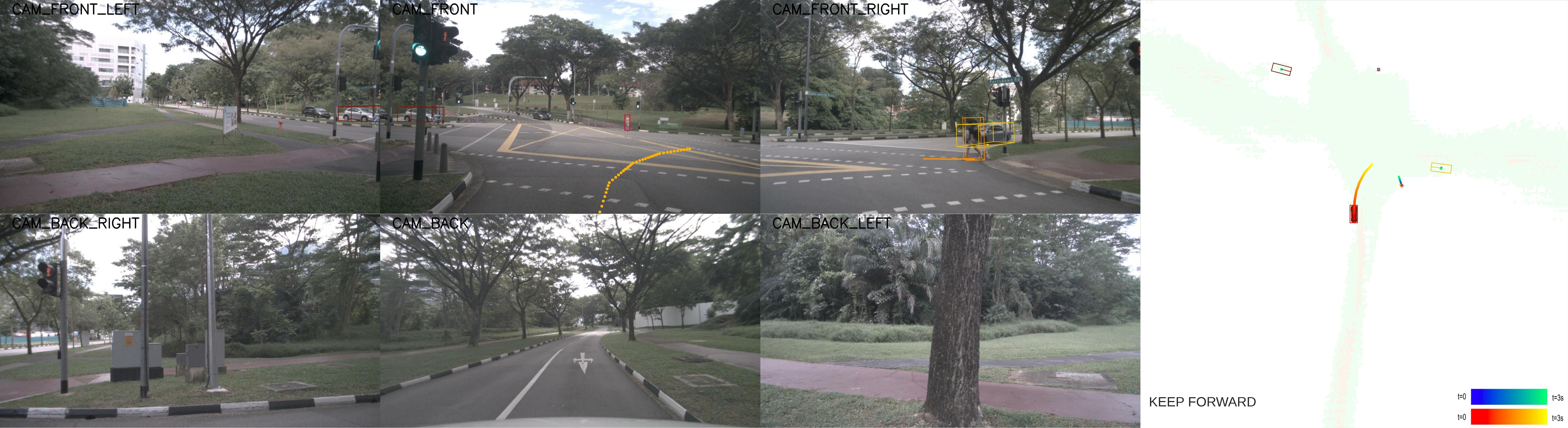}\\
    \small (b) OpenDriveVLA planning under instruction turning right

    \vspace{1.2em}

    % Bottom image
    \includegraphics[width=0.98\linewidth]{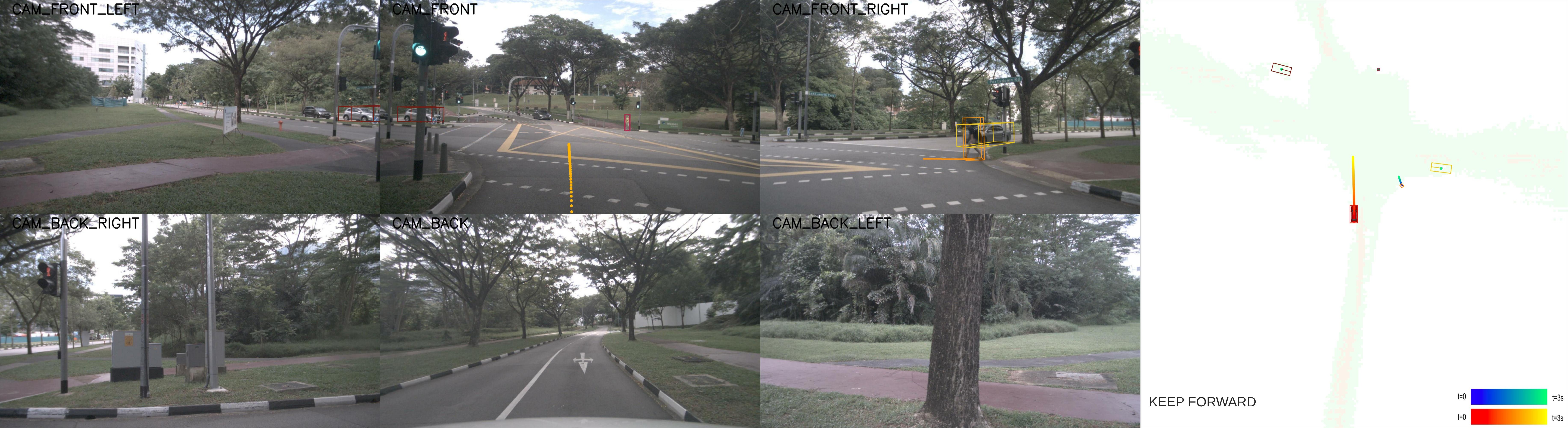}\\
    \small (c) OpenDriveVLA planning under ground-truth instruction: keep forward
    
    \caption{Example 2 of trajectory planning results of OpenDriveVLA after stage 3 with different driving instructions.}
    \label{fig:stage3_compare2}
\end{figure}

\clearpage
\subsection{Discussions}
\subsubsection{Results Regarding Model Size.}
The results in Main Content (Table 1,2,3) show that, while increasing model size generally leads to improved performance, the larger OpenDriveVLA-7B model does not consistently outperform the smaller 0.5B and 3B variants across all benchmarks. Specifically, on the Nu-X dataset, the 0.5B model achieves obviously higher CIDEr and ROUGE-L scores than the 7B version. In open-loop trajectory planning, the differences in average L2 error and collision rate among the three models are relatively small. Although the 7B model remains competitive overall, its performance gains are not always significant and, in some cases, fall behind those of the smaller models.

There are several possible reasons based on our experimental observations. First, the performance gains associated with scaling large language models typically depend on access to extensive and diverse training data. As detailed in Table~\ref{tab:qa_dataset_details}, the instruction tuning data and domain-specific driving annotations used in our current setup may not be sufficient to fully leverage the representation capacity of the 7B model, especially in the autonomous driving domain, where large-scale open-source vision language datasets are still very few. Hence, the expected improvements from model scaling are not consistently realized across all tasks.

Second, larger autoregressive models tend to rely more heavily on language priors during generation. In structured multimodal reasoning tasks, this reliance may weaken the model’s ability to maintain accurate visual grounding, particularly in scenarios requiring fine-grained spatial understanding or factual precision. Smaller models, though with limited capacity, can retain stronger coupling between vision and language inputs, which can lead to more stable behavior under constrained supervision. This also suggests that additional high-quality visual-textual data is necessary to facilitate further development.

Third, the optimization dynamics of larger models can be more sensitive to hyperparameters and experimental settings in domain-specific tasks. Without sufficient  diversity in the training dataset, the larger model may overfit to dominant patterns in the training set, resulting in reduced robustness and generalization during inference.

In summary, model size alone is not a robust indicator of performance in this domain-specific autonomous driving task with limited data sources. Instead, data quantity and quality, and multi-stage training strategies play a critical role in achieving robust and scalable performance.

\subsubsection{Current Limitations}
While OpenDriveVLA achieves promising results, several limitations remain. First, the model relies on implicit reasoning patterns acquired through instruction tuning, without explicit step-by-step chain-of-thought deduction during inference. Although this design helps maintain inference efficiency, it may eteriorate the reasoning performance and limit the model’s ability to handle complex scenarios. Second, the autoregressive nature of LLM introduces inference latency. Even with careful control of token lengths, the sequential decoding process remains a bottleneck for deployment in high-speed driving scenarios. Further improvements in model quantization and improved decoding strategies will be beneficial to meet real-time requirements. Finally, our current planning evaluation is limited to an open-loop setting. This does not capture the interactive feedback dynamics of realistic driving and may lead to overestimated robustness, underscoring the importance of extending OpenDriveVLA to closed-loop, language-aware simulation environments.

\subsubsection{Extension to Closed-Loop Planning}

As mentioned, the current planning evaluation of OpenDriveVLA is conducted in an open-loop setting on the nuScenes dataset. While this setup allows for controlled and reproducible benchmarking, it does not account for the feedback dynamics of ego-agent interactions in real-world traffic. Prior studies also show that nuscences open-loop benchmark may lead to overly optimistic conclusions\cite{ego-mlp}. Several established closed-loop benchmarks, such as nuPlan\cite{caesar2024nuplan}, Bench2Drive\cite{jia2024bench2drive}, and NaviSim\cite{dauner2024navsim}, have been introduced for more reliable planning assessment.

Unfortunately, these benchmarks currently lack vision-language annotations that are central to the training of domain-specific LLM-based Autonomous Driving models. While nuScenes-derived datasets such as TOD3Cap, nuScenes QA, nuCaption, and nuX provide rich multimodal supervision in the form of instance-level and scene-level captions and question–answer pairs. These additional datasets are the result of extensive annotation efforts from multiple prior works and serve as a foundation for both training and evaluating LLM-based autonomous driving models. To extend OpenDriveVLA toward closed-loop evaluation, future work involves building an automated data generation pipeline that (semi-)automatically generates instance-level, map-level, and scene-level descriptions, as well as question–answer pairs linked to driving behaviors. 

Despite the limitations, OpenDriveVLA still demonstrates strong performance across both open-loop planning and multiple driving VQA benchmarks, including nuScenes-QA, nuCaption, and nuX. The model consistently exhibits robust scene understanding, accurate instruction following, and semantically grounded action generation. These results validate the effectiveness of the proposed model and training techniques, and provide a solid experimental foundation for future extensions into closed-loop evaluation frameworks.

% limitation？
% ===============================================================

% ======== 目录在最后生成 ========
\clearpage
% \renewcommand{\contentsname}{Appendix Contents}
% \tableofcontents
% ==============================================

\endgroup
%%%%%%%%%%%%%%%%%%%%%%%%%%%%%%%%%%%%%%%%%%%%%%%%%%%%%%%%%%%%%%%%%%%%%%

%\end{sloppypar}
\end{document}